\documentclass[11pt]{article}

\usepackage[final]{acl}

\usepackage{times}
\usepackage{latexsym}
\usepackage[T1]{fontenc}
\usepackage[utf8]{inputenc}
\usepackage{microtype}
\usepackage{inconsolata}
\usepackage{graphicx}
\usepackage{xcolor}
\usepackage{url}
\usepackage{amsmath,amssymb}
\usepackage{graphicx}
\usepackage{enumitem}
\usepackage{booktabs}
\usepackage{tabularx}
\usepackage{makecell}
\usepackage{multirow}
\usepackage{subcaption}
\usepackage{wrapfig}
\usepackage{float}
\usepackage[most]{tcolorbox}

\definecolor{mydarkblue}{rgb}{0.0, 0.0, 0.5}
\newcommand{\refig}[1]{Fig.~\ref{#1}}
\newcommand{\refapp}[1]{App.~\ref{#1}}

\title{SafeMERGE: Preserving Safety Alignment in Fine-Tuned Large Language Models via Selective Layer-Wise Model Merging}

\author{%
  Aladin Djuhera$^1$, Swanand Ravindra Kadhe$^2$, Farhan Ahmed$^2$, Syed Zawad$^2$, Holger Boche$^1$ \\
  $^1$ Technical University Munich $^2$ IBM Research
}

\begin{document}

\maketitle


\begin{abstract}
    Fine-tuning large language models (LLMs) is a common practice to adapt generalist models to specialized domains.
    However, recent studies show that fine-tuning can erode safety alignment, causing LLMs to respond to harmful or unethical prompts.
    Many methods to realign safety have been proposed, but often introduce custom algorithms that are difficult to implement or compromise task utility.
    In this work, we propose SafeMERGE\footnote{Code available at: \href{https://github.com/aladinD/SafeMERGE}{github.com/aladinD/SafeMERGE}}, a lightweight, \emph{post-fine-tuning} framework that restores safety while maintaining downstream performance.
    SafeMERGE selectively merges fine-tuned with safety-aligned model layers \emph{only} when they deviate from safe behavior, measured by a cosine similarity criterion.
    Across four LLMs and several tasks, SafeMERGE consistently reduces harmful outputs compared to other defenses, with negligible or even positive impact on utility.
    Our results demonstrate that selective, layer-wise merging offers a robust safeguard against the inadvertent loss of safety during fine-tuning, establishing SafeMERGE as a simple yet effective post-fine-tuning defense.
\end{abstract}


\section{Introduction and Motivation}
\label{sec:introduction}

\begin{figure}[t]
    \centering
    \includegraphics[width=0.45\textwidth]{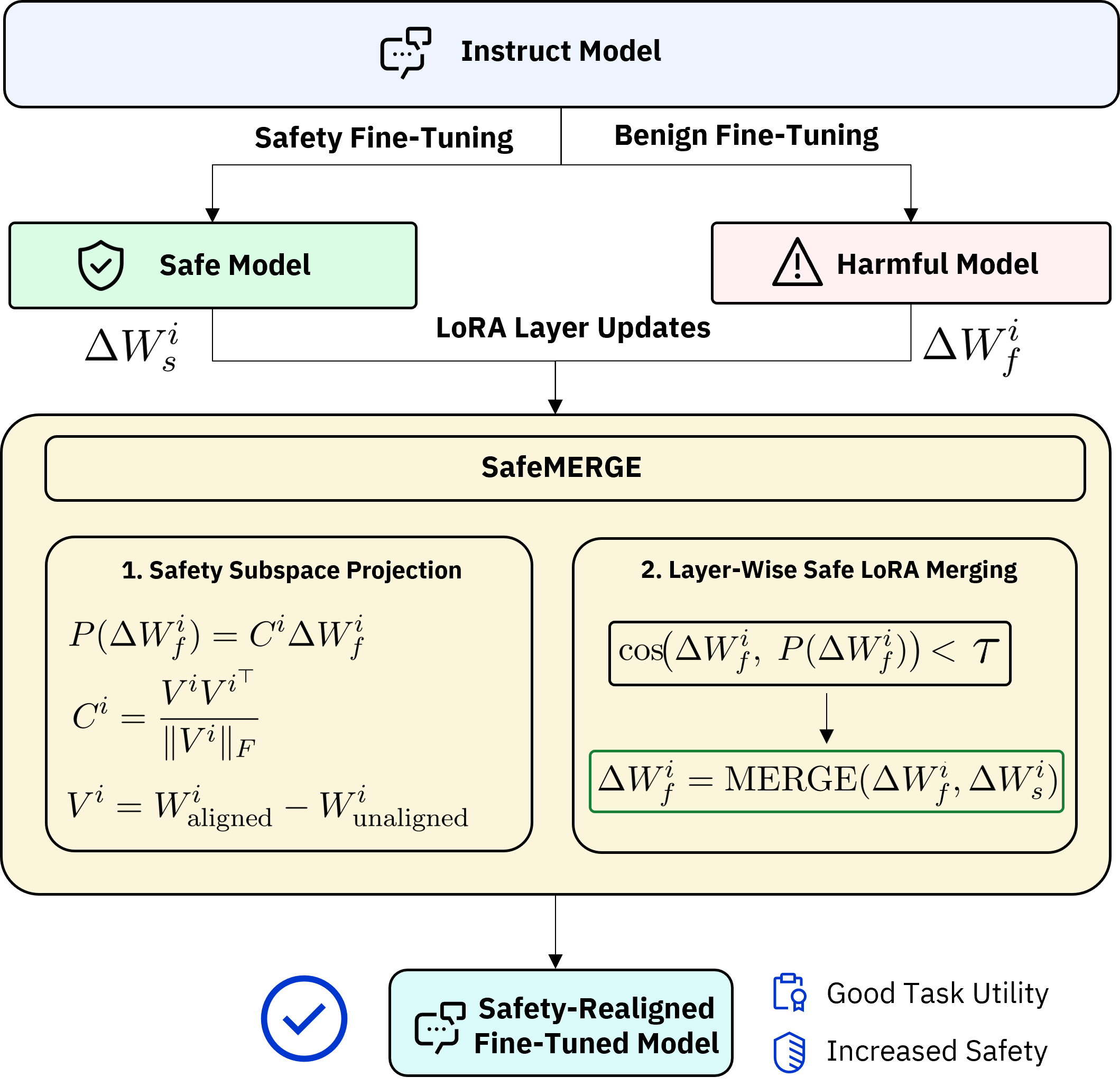}
    \caption{SafeMERGE merges harmful and safe LoRA adapters only if the layers deviate from safe behavior, measured by a projection-based cosine similarity.}
    \label{fig:safemerge_diagram}
    \vspace{-10pt}
\end{figure}

Large language models (LLMs) have demonstrated remarkable capabilities while becoming increasingly accessible.
Practitioners typically adapt these models to specialized domains, such as code and math, by fine-tuning with domain-specific data.
In this process, safety tuning is critical to ensure that LLMs remain aligned with human values and security policies \citep{ouyang2022training, bai2022constitutional, zhang2024instructiontuninglargelanguage}.
However, safety alignment has been shown to be fragile during various stages of adaptation \citep{wei2023jailbroken, huang2024catastrophic, zeng2024johnny, zhan2024removing}.
For instance, \citet{yang2023shadow} demonstrate that fine-tuning with only a few malicious training examples can jailbreak LLMs, prompting them to comply with harmful user requests.
More concerningly, \citet{anonymous2024finetuning} show that even benign fine-tuning can inadvertently degrade safety.
Theoretical works on refusal directions \citep{arditi2024refusallanguagemodelsmediated} and token-depth \citep{qi2024safetyalignmentjusttokens} further suggest that safety alignment is often merely shallow and easily broken.
Ensuring that LLMs \emph{remain safe after fine-tuning} is therefore a critical practical challenge.

Recent defenses that address this challenge can be broadly categorized into three groups based on the stage at which interventions are applied:
(a) \emph{alignment-stage defenses}, which intervene during the initial safety alignment process \citep{huang2024vaccineperturbationawarealignmentlarge, rosati2024representationnoisingdefencemechanism},
(b) \emph{fine-tuning-stage defenses}, which modify the training procedure during domain adaptation \citep{bianchi2024safetytunedllamaslessonsimproving, qi2024safetyalignmentjusttokens, huang2024lisa}, and
(c) \emph{post-fine-tuning-stage defenses}, which attempt to restore safety after fine-tuning has completed \citep{bhardwaj2024languagemodelshomersimpson, hsu2025safelorasilverlining}.
An overview of corresponding recent works is provided in \refapp{sec:A_related_work}.

However, many defenses rely on custom alignment or complex fine-tuning algorithms that are difficult to integrate with standard open-source libraries and demand specialized expertise, thereby hindering practical adoption.
Moreover, simpler defenses that avoid such custom training often compromise task performance in favor of safety.
Motivated by these practical challenges, we ask:
\emph{How can practitioners retain task utility while improving safety, without relying on custom algorithms, and while using standard open-source libraries?}

In this paper, we propose \textbf{SafeMERGE}, a lightweight, post-fine-tuning framework that \emph{selectively merges} fine-tuned model layers with those from a safety-aligned model, but \emph{only} when they deviate from safe behavior.
To identify such deviations, SafeMERGE measures the cosine similarity between layer activations and their projection onto a safety-aligned subspace, and merges only those layers that fall outside a similarity threshold.
\refig{fig:safemerge_diagram} illustrates the general approach.
With SafeMERGE, we specifically address practitioners who aim to restore safety after fine-tuning \emph{without requiring changes to their existing training pipeline}.
The result is a simple but principled framework that integrates easily into standard workflows.

We evaluate SafeMERGE on four widely used LLMs: Llama-2-7B-Chat \citep{touvron2023llama2openfoundation}, Llama-3.1-8B-Instruct \citep{grattafiori2024llama3}, Qwen-2-7B-Instruct \citep{yang2024qwen2technicalreport}, and Qwen-2.5-7B-Instruct \citep{qwen2025qwen25technicalreport}.
We fine-tune each model on two standard tasks in the main body, GSM8K \citep{cobbe2021gsm8k} and PubMedQA \citep{jin2019pubmedqa}, and further evaluate three out-of-distribution telecom tasks in the appendix.
In our experiments, we demonstrate that SafeMERGE significantly reduces harmfulness while maintaining strong task performance, thereby achieving a better trade-off between utility and safety compared to existing baselines that similarly operate without custom fine-tuning algorithms.
We further conduct several ablation studies to investigate key components, including different merging strategies, weighting schemes, and similarity thresholds.
\section{SafeMERGE: Selective Layer-Wise Safe LoRA Model Merging}
\label{sec:safemerge}

Given an aligned model (e.g., an instruct or chat variant) and a task-specific dataset, our goal is to fine-tune the model to maximize task utility while minimizing safety degradation.
To this end, we focus on parameter-efficient LoRA fine-tuning~\citep{hu2021loralowrankadaptationlarge}, which is widely adopted in practice (see \refapp{sec:lora}).
SafeMERGE achieves this goal by constructing two complementary models: (i) the fine-tuned model trained on task-specific data, and (ii) a safe model trained on safety-aligned data (e.g., harmful-prompt–safe-response pairs).
It then selectively merges only those fine-tuned layers that underwent safety degradation with the corresponding layers from the safe model.
While this introduces an additional preparation step, our experiments show that fine-tuning the safety model is straightforward and requires only little data that can be sourced from publicly available safety datasets (see \refapp{sec:B_2}).
Furthermore, because the safe model is \emph{task-agnostic}, it only needs to be trained once and can be reused across different fine-tuning tasks.

Inspired by \citet{hsu2025safelorasilverlining}, SafeMERGE identifies unsafe layers by constructing a \emph{safety-aligned subspace} \( V^i \) and measuring each layer's cosine similarity to it.
Specifically, $V^i$ is computed as the difference between the weights of the aligned (e.g., instruct) and unaligned (e.g., base) models, i.e.,
\begin{equation}
    V^i = W_{\mathrm{aligned}}^i - W_{\mathrm{unaligned}}^i .
\end{equation}
This subspace represents the safety alignment in the weight space per layer, and the projection $C^i$ onto it can be computed as \( C^i = \frac{V^i V^{i^\top}}{\|V^i\|_F} \).

Consequently, a smaller cosine similarity between fine-tuned (\(\Delta W_f^i\)) and projected (\(C^i \Delta W_f^i\)) LoRA layers indicates a greater deviation from the safety-aligned subspace. 
This allows us to identify harmful layers as follows. 
Let \( \rho^i \) denote the cosine similarity between \(\Delta W_f^i\) and \(C^i \Delta W_f^i\), i.e., 
\begin{equation}
    \rho^i = \cos\left(\Delta W_f^i, C^i \Delta W_f^i\right) .
\end{equation}
Given a safety threshold \(\tau \in (0, 1)\), the fine-tuned layer \(\Delta W_f^i\) is considered \emph{unsafe} if \(\rho^i < \tau\). 
For each such layer, SafeMERGE merges it with the corresponding safe model layer \(\Delta W_s^i\), i.e.,
\begin{equation}
    \Delta W_{\mathrm{merge}}^i = \mathrm{MERGE}(\Delta W_f^i, \Delta W_s^i) ,
\end{equation}
where \(\mathrm{MERGE}(\cdot)\) defines the merging strategy. 
One example is linear merging with \(\alpha \in [0,1]\) \cite{ilharco2023editing}, i.e.,
\begin{equation}
\Delta W_{\mathrm{merge, linear}}^i = \alpha \Delta W_f^i + (1 - \alpha)\,\Delta W_s^i .
\label{eq:linear_merging}
\end{equation}

Note that the threshold \(\tau\) controls the selectivity, where a larger \(\tau\) implies merging more safe layers, while a smaller \(\tau\) retains more fine-tuned updates.

We argue that SafeMERGE's \emph{selective merging strategy} yields a more favorable safety--utility trade-off compared to purely projection-based approaches such as SafeLoRA~\citep{hsu2025safelorasilverlining}, and methods that indiscriminately merge model layers \citep{bhardwaj2024languagemodelshomersimpson, farn2025safeguardfinetunedllmspre}.
Our empirical results support this claim, demonstrating that SafeMERGE consistently outperforms other baselines across multiple settings while remaining lightweight, as it requires no retraining and can thus be run entirely on CPU (see \refapp{sec:computational_analysis}).
\section{Experimental Setup}
\label{sec:experimental_setup}

\paragraph{Models and Datasets.}
We LoRA fine-tune four widely used LLMs: Llama-2-7B-Chat \citep{touvron2023llama2openfoundation}, Llama-3.1-8B-Instruct \citep{grattafiori2024llama3}, Qwen-2-7B-Instruct \citep{yang2024qwen2technicalreport}, and Qwen-2.5-7B-Instruct \citep{qwen2025qwen25technicalreport}.
Our \emph{primary utility datasets} are GSM8K \citep{cobbe2021gsm8k}, a math corpus with grade-school problems for multi-step reasoning, and PubMedQA \citep{jin2019pubmedqa}, a biomedical corpus with more samples and a broader domain shift.
To further assess generalizability to highly specialized domains, in \refapp{sec:telecom_results}, we additionally fine-tune the models on three out-of-distribution datasets from the telecom domain: TeleData \citep{TeleLLMs}, TeleQnA \citep{TeleQnA}, and TSpecLLM \citep{TSpecLLM}.
These contain various questions drawn from telecom standards and engineering practice, often formatted as lists, tables, and complex mathematical formulas, which are shown to be harmful during training \citep{he2024safedataidentifyingbenign, djuhera2026safecommstudysafetydegradation}.
Further details on utility fine-tuning are provided in \refapp{sec:utility_fine_tuning}.

\paragraph{Evaluation Setup.}
To assess task performance on the \emph{utility datasets}, we report exact-match accuracy for GSM8K and classification accuracy for PubMedQA. 
We also assess \emph{general abilities} via IFEval \citep{zhou2023instructionfollowingevaluationlargelanguage} and MMLU \citep{hendryckstest2021}.
For \emph{safety evaluations}, we follow \citet{yao2024survey, qi2024safetyalignmentjusttokens, hsu2025safelorasilverlining} and generate responses on DirectHarm \citep{lyu2024keeping} and HexPhi \citep{anonymous2024finetuning}, two popular red-teaming benchmarks.
We use Llama-Guard-3-8B \citep{grattafiori2024llama3} to assess safety and report the overall harmfulness score as the proportion of model responses flagged as unsafe. 
To mitigate evaluator bias, in \refapp{sec:safety_guard_model}, we cross-validate our harmfulness scores using ShieldGemma-9B \citep{shieldgemma}, observing consistent trends with the Llama-Guard model.
More details on our evaluation setup are provided in \refapp{sec:C_evaluation_setup}.

\paragraph{Baselines.}
We compare SafeMERGE against methods that similarly require no custom alignment:
\emph{a) SafeInstruct}~\citep{bianchi2024safetytunedllamaslessonsimproving}, a fine-tuning defense that augments the training data with additional safety samples,
\emph{b) RESTA}~\citep{bhardwaj2024languagemodelshomersimpson}, a post-fine-tuning method that removes harmful task vectors by subtracting parameters of an unsafe model from the fine-tuned one,
\emph{c) RESTA-Instruct}~\citep{farn2025safeguardfinetunedllmspre}, a RESTA variant that instead performs full-parameter merging with instruct models to induce safe task vectors, and
\emph{d) SafeLoRA}~\citep{hsu2025safelorasilverlining}, which projects LoRA updates onto a safety-aligned subspace.
Further background, discussions, and intermediate results are provided in \refapp{sec:D_baselines}.

\begin{table*}[t]
    \centering
    \caption{SafeMERGE compared to baselines (SafeInstruct, RESTA, RESTA-Instruct, SafeLoRA) on Llama and Qwen models fine-tuned on GSM8K and PubMedQA. Utility is measured on the target task, general capabilities on IFEval/MMLU, and harmfulness via DirectHarm/HexPhi. Best results are \textbf{bolded} and second-best are \underline{underlined}.}
    \label{tab:consolidated_results}
    \resizebox{1\linewidth}{!}{
    \begin{tabular}{llccccccc}
        \toprule
        \textbf{Model} & \textbf{Benchmark} & \textbf{Original} & \textbf{Fine-tuned} & \textbf{SafeInstruct} & \textbf{RESTA} & \textbf{RESTA-Instruct} & \textbf{SafeLoRA} & \textbf{SafeMERGE} \\
        \midrule

        \toprule
        \multirow{5}{*}{\begin{tabular}{@{}c@{}}Llama-2-7B-Chat \\ (GSM8K)\end{tabular}}
          & GSM8K $(\uparrow)$         & 22.67 & \textbf{27.37} & 26.00 & 24.94 & 25.90 & 26.15 & \underline{26.96} \\
          & IFEval $(\uparrow)$        & \textbf{40.30} & 40.00 & \underline{40.10} & 39.70 & 39.90 & 40.00 & \textbf{40.30} \\
          & MMLU $(\uparrow)$          & \textbf{23.60} & 23.40 & \underline{23.50} & 23.00 & 23.20 & 23.40 & \underline{23.50} \\
          & DirectHarm $(\downarrow)$  & \textbf{5.00}  & 27.80 & \underline{7.50}  & \underline{7.50}  & 9.50  & 10.20 & \underline{7.50} \\
          & HexPhi $(\downarrow)$      & \textbf{2.00}  & 16.40 & 6.20  & \underline{4.30}  & 6.80  & 6.90  & 5.70 \\
        \midrule

        \multirow{5}{*}{\begin{tabular}{@{}c@{}}Llama-2-7B-Chat \\ (PubMedQA)\end{tabular}}
          & PubMedQA $(\uparrow)$      & 55.20 & \textbf{72.60} & 71.20 & 57.10 & 64.50 & 71.40 & \underline{72.20} \\
          & IFEval $(\uparrow)$        & \underline{40.30} & 40.00 & 40.10 & 39.60 & 39.80 & 40.00 & \textbf{40.40} \\
          & MMLU $(\uparrow)$          & \textbf{23.60} & 23.30 & \underline{23.40} & 22.90 & 23.10 & 23.30 & \underline{23.40} \\
          & DirectHarm $(\downarrow)$  & \textbf{5.00}  & 12.50 & 12.20 & \underline{5.80}  & 8.10  & 10.70 & 8.10 \\
          & HexPhi $(\downarrow)$      & \textbf{2.00}  & 6.20  & 6.30  & \underline{4.20}  & 5.30  & 5.90  & 4.30 \\

        \toprule
        \multirow{5}{*}{\begin{tabular}{@{}c@{}}Llama-3.1-8B-Instruct \\ (GSM8K)\end{tabular}}
          & GSM8K $(\uparrow)$         & 73.80 & \underline{78.24} & 77.40 & 74.20 & 77.10 & 77.90 & \textbf{78.50} \\
          & IFEval $(\uparrow)$        & \textbf{78.30} & 78.10 & \underline{78.20} & 77.40 & 77.90 & 78.00 & \textbf{78.30} \\
          & MMLU $(\uparrow)$          & 63.10 & 62.80 & \underline{62.90} & 61.70 & 62.30 & 62.70 & \textbf{63.20} \\
          & DirectHarm $(\downarrow)$  & \underline{11.30} & 28.30 & 12.50 & 11.90 & 13.50 & 15.10 & \textbf{8.80} \\
          & HexPhi $(\downarrow)$      & 7.90  & 14.70 & 7.20  & \underline{6.90}  & 7.20  & 7.10  & \textbf{6.30} \\
        \midrule

        \multirow{5}{*}{\begin{tabular}{@{}c@{}}Llama-3.1-8B-Instruct \\ (PubMedQA)\end{tabular}}
          & PubMedQA $(\uparrow)$      & 74.40 & \underline{78.80} & 78.50 & 75.70 & 77.40 & 78.30 & \textbf{79.00} \\
          & IFEval $(\uparrow)$        & \textbf{78.30} & 78.00 & 78.10 & 77.30 & 77.80 & 78.00 & \underline{78.20} \\
          & MMLU $(\uparrow)$          & \textbf{63.10} & 62.70 & \underline{62.80} & 61.50 & 62.20 & 62.60 & \underline{62.80} \\
          & DirectHarm $(\downarrow)$  & 11.30 & 23.50 & 11.80 & \underline{10.30} & 14.20 & 16.70 & \textbf{9.10} \\
          & HexPhi $(\downarrow)$      & 7.90  & 12.20 & 9.70  & \underline{7.10}  & 8.70  & 9.60  & \textbf{6.80} \\

        \toprule
        \multirow{5}{*}{\begin{tabular}{@{}c@{}}Qwen-2-7B-Instruct \\ (GSM8K)\end{tabular}}
          & GSM8K $(\uparrow)$         & 58.38 & 70.13 & 72.69 & 60.73 & 69.30 & \textbf{74.37} & \underline{72.90} \\
          & IFEval $(\uparrow)$        & \textbf{56.00} & 55.80 & \underline{55.90} & 54.70 & 55.40 & 55.80 & \underline{55.90} \\
          & MMLU $(\uparrow)$          & \textbf{69.00} & 68.70 & \underline{68.90} & 67.90 & 68.30 & 68.60 & 68.80 \\
          & DirectHarm $(\downarrow)$  & 18.20 & 25.30 & \underline{13.70} & 18.80 & 17.40 & 22.30 & \textbf{8.20} \\
          & HexPhi $(\downarrow)$      & 11.50 & 16.80 & \underline{9.50}  & 15.80 & 14.10 & 14.80 & \textbf{7.50} \\
        \midrule

        \multirow{5}{*}{\begin{tabular}{@{}c@{}}Qwen-2-7B-Instruct \\ (PubMedQA)\end{tabular}}
          & PubMedQA $(\uparrow)$      & 73.60 & 79.60 & 80.00 & 75.80 & 78.50 & \textbf{82.80} & \underline{80.30} \\
          & IFEval $(\uparrow)$        & \textbf{56.00} & 55.70 & \underline{55.80} & 54.60 & 55.30 & 55.70 & \underline{55.80} \\
          & MMLU $(\uparrow)$          & \textbf{69.00} & 68.60 & 68.80 & 67.80 & 68.20 & 68.50 & \underline{68.90} \\
          & DirectHarm $(\downarrow)$  & 18.20 & 26.00 & \underline{12.50} & 18.50 & 17.60 & 19.50 & \textbf{8.50} \\
          & HexPhi $(\downarrow)$      & \underline{11.50} & 13.20 & \textbf{5.90} & 14.80 & 14.50 & 14.50 & \textbf{5.90} \\

        \toprule
        \multirow{5}{*}{\begin{tabular}{@{}c@{}}Qwen-2.5-7B-Instruct \\ (GSM8K)\end{tabular}}
          & GSM8K $(\uparrow)$         & 54.60 & 77.00 & 77.80 & 56.30 & 65.70 & \underline{78.10} & \textbf{78.40} \\
          & IFEval $(\uparrow)$        & \textbf{72.40} & 71.80 & 71.80 & 71.00 & 71.30 & 71.70 & \underline{71.90} \\
          & MMLU $(\uparrow)$          & \underline{68.70} & 68.50 & \underline{68.70} & 67.70 & 68.30 & 68.40 & \textbf{68.80} \\
          & DirectHarm $(\downarrow)$  & 14.20 & 22.10 & \underline{12.50} & 18.30 & 17.70 & 19.40 & \textbf{10.10} \\
          & HexPhi $(\downarrow)$      & 13.20 & 17.30 & \underline{11.30} & 16.20 & 15.40 & 14.40 & \textbf{10.30} \\
        \midrule

        \multirow{5}{*}{\begin{tabular}{@{}c@{}}Qwen-2.5-7B-Instruct \\ (PubMedQA)\end{tabular}}
          & PubMedQA $(\uparrow)$      & 73.20 & 78.80 & \underline{79.20} & 74.20 & 77.60 & 79.10 & \textbf{79.50} \\
          & IFEval $(\uparrow)$        & \textbf{72.40} & 72.10 & 72.00 & 71.20 & 71.50 & 71.90 & \underline{72.30} \\
          & MMLU $(\uparrow)$          & \underline{68.70} & 68.40 & \underline{68.70} & 67.80 & 68.40 & 68.60 & \textbf{68.90} \\
          & DirectHarm $(\downarrow)$  & 14.20 & 20.20 & \underline{13.10} & 15.70 & 14.90 & 17.10 & \textbf{11.10} \\
          & HexPhi $(\downarrow)$      & 13.20 & 17.80 & \underline{9.50} & 15.60 & 15.20 & 13.30 & \textbf{8.70} \\

        \bottomrule
    \end{tabular}
    }
\end{table*}

\paragraph{SafeMERGE.}
We selectively merge harmful fine-tuned with safety-aligned LoRA layers \emph{only} where the former fail the cosine similarity test.
To this end, we obtain the safe model by fine-tuning on subsets (100, 500, 1000, 2500 samples) of the public safety dataset from \citet{bianchi2024safetytunedllamaslessonsimproving} and selecting the model with the lowest harmfulness (see \refapp{sec:B_2}).
Similar to \citet{hsu2025safelorasilverlining}, we use base and chat/instruct models to define the safety-aligned subspace, allowing for a direct comparison.
However, other models, such as explicitly safety-tuned ones, can also be used (see \refapp{sec:safelora_implementation}).
We investigate different weighting schemes (see \refapp{sec:E_3}) and additionally explore DARE \citep{DARE} and TIES \citep{TIES} merging strategies, but find standard linear merging sufficient (see \refapp{sec:E_4}).
\section{Results and Discussion}
\label{sec:results}

We compare SafeMERGE to all baselines with a focus on \emph{linear merging} and analyze key performance trends. 
Table \ref{tab:consolidated_results} reports the main results for the GSM8K and PubMedQA tasks.

\paragraph{Overall Performance.}
In general, SafeMERGE matches or exceeds utility while significantly reducing harmfulness.
For Llama-2 (GSM8K), it retains near-best accuracy (26.96\%) while reducing Direct-Harm (HexPhi) from 27.80\% (16.40\%) to 7.50\% (5.70\%).
For Llama-3.1 (GSM8K), SafeMERGE even improves accuracy to 78.50\%, surpassing the fine-tuned model while achieving the lowest harmfulness, even lower than the original instruct model.
Similar trends hold for Llama-2 and Llama-3.1 on PubMedQA, and for Qwen-2 and Qwen-2.5 models on both datasets. 
SafeMERGE achieves this with selective merging: only 28 LoRA layers for Llama-2, 29 for Llama-3.1, and 34 for Qwen-2/2.5.
Similar trends can be observed for the telecom tasks (see Table \ref{tab:results_telecom} in \refapp{sec:extended_results}), where SafeMERGE consistently restores safety while retaining high utility.

\paragraph{Baseline Comparisons.}
Overall, SafeMERGE delivers the best safety--utility trade-off across all models and tasks.
Among baselines, SafeInstruct is the strongest competitor, whereas RESTA and RESTA-Instruct consistently underperform on utility.
SafeLoRA ranks second in utility (after SafeMERGE) but falls short in safety, lagging behind the other methods.
Results on IFEval and MMLU further show that all defenses preserve general reasoning.
This aligns with prior findings \citep{arditi2024refusallanguagemodelsmediated, hsu2025safelorasilverlining}, confirming that safety interventions primarily regulate refusal behavior without degrading broader reasoning or knowledge capabilities.
These findings demonstrate that SafeMERGE's selective merging a) avoids the utility degradation of indiscriminate merging (RESTA variants), b) restores safety better than projection-based alignment (SafeLoRA), and c) surpasses even fine-tuning-stage methods (SafeInstruct).
\section{Insights from Ablations}
\label{sec:insights}

\begin{table}[t]
\centering
\caption{Cosine similarity vs.\ Euclidean distance as intervention metrics on GSM8K.}
\label{tab:cosine_vs_euclidean}
\resizebox{1\linewidth}{!}{%
\begin{tabular}{llcc}
\toprule
\textbf{Fine-Tuned Model} & \textbf{Intervention Metric} & \textbf{GSM8K $(\uparrow)$} & \textbf{DirectHarm $(\downarrow)$} \\
\midrule
\multirow{3}{*}{\begin{tabular}{@{}c@{}}Llama-3.1-8B-Instruct \\ (GSM8K)\end{tabular}}
& None (Baseline)       & 78.24 & 28.30 \\
& Euclidean Distance       & 42.10 & 17.40 \\
& \textbf{Cosine Similarity} & \textbf{78.50} & \textbf{8.80} \\
\midrule
\multirow{3}{*}{\begin{tabular}{@{}c@{}}Qwen-2-7B-Instruct \\ (GSM8K)\end{tabular}}
& None (Baseline)       & 70.13 & 25.30 \\
& Euclidean Distance       & 45.30 & 15.80 \\
& \textbf{Cosine Similarity} & \textbf{72.90} & \textbf{8.20} \\
\bottomrule
\end{tabular}
}
\end{table}

We summarize key ablation results.
Additional analyses are provided in \refapp{sec:E_results} and \refapp{sec:extended_results}.

\paragraph{Cosine Similarity as an Intervention Metric.}

SafeMERGE adopts the cosine similarity criterion because it captures the directional alignment between fine-tuned updates and the safety-aligned subspace, independent of update magnitude.
This is particularly important in high-dimensional weight spaces, where fine-tuning may induce large parameter shifts without necessarily compromising safety.
In contrast, magnitude-based metrics, such as Euclidean distance, conflate \emph{how much} a model has changed with \emph{whether} it has drifted in an unsafe direction, and may therefore flag useful task updates as harmful even when safe.
To validate our design choice, we compare cosine similarity against Euclidean distance on GSM8K for the Llama-3.1-8B and Qwen-2-7B instruct models in Table~\ref{tab:cosine_vs_euclidean}.
For Llama, using Euclidean distance for intervention reduces harmfulness on Direct-Harm to 17.40\%, but causes a severe utility drop to 42.10\%.
In contrast, cosine-similarity-based SafeMERGE improves utility to 78.50\% while reducing harmfulness much more substantially to 8.80\%.
A similar pattern holds for the Qwen model, which sees a significantly better safety--utility trade-off when using cosine similarity as the intervention metric. 
These results confirm that cosine similarity is a substantially more reliable proxy for safety degradation, as it restores safety without sacrificing the fine-tuned model's downstream capabilities.

\paragraph{Mechanistic Interpretability of Threshold \texorpdfstring{$\tau$}{tau}.}

\begin{figure}[t]
    \centering
    \includegraphics[width=1\linewidth]{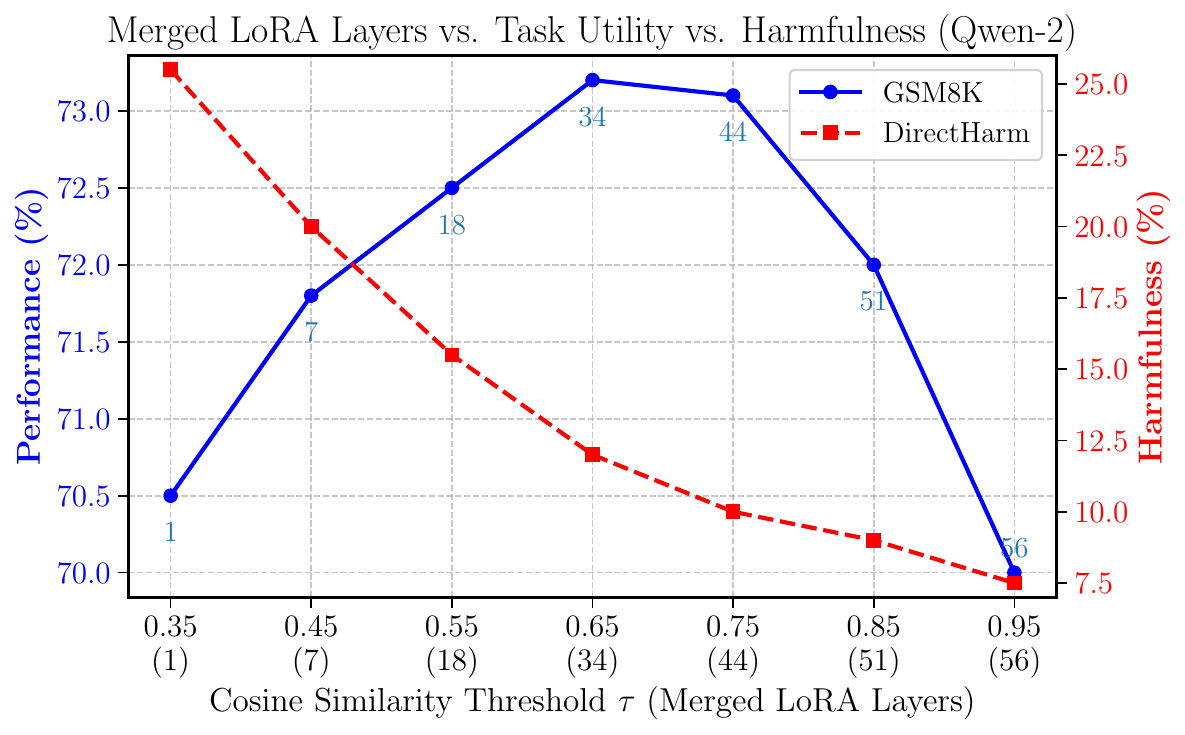}
    \caption{Qwen-2-7B-Instruct (GSM8K, DirectHarm) with weighting factor \( \alpha = 0.7 \). The optimal trade-off between utility and safety is achieved for \( \tau = 0.65 \). Similar results and patterns are observed for Qwen-2.5.}
    \label{fig:safemerge_threshold_analysis_qwen2_gsm8k_main}
\end{figure}

Because cosine similarity measures the angular alignment between a fine-tuned LoRA update and the safety-aligned direction, the choice of \(\tau\) directly determines how far a layer may deviate from the safety manifold before intervention is triggered.
This makes \(\tau\) geometrically interpretable: layers are merged only when their updates become sufficiently orthogonal or opposed to the safety direction.
Further, our ablations identify \emph{stable heuristics and reliable defaults} that provide strong safety--utility trade-offs across settings, reducing the need for granular per-task fine-tuning.
Specifically, optimal thresholds remain within a narrow range across models and datasets, typically around \(\tau \in [0.6, 0.8]\), with representative optima \(0.7\) for Llama-2, \(0.75\) for Llama-3.1, and \(0.65\) for Qwen-2 across tasks (see \refig{fig:safemerge_threshold_analysis_qwen2_gsm8k_main} and \refapp{sec:E_2}).

\paragraph{Weighting and Merging Schemes.}

As the similarity threshold \(\tau\) increases, more layers are merged, progressively improving safety at the cost of task utility.
Weighting schemes further shape this trade-off by controlling the relative contribution of the models.
Across experiments, balanced weights that sum to 1.0 generally outperform overweighted ratios with disproportionate contributions (see \refapp{sec:E_3}).
We also find that alternative merging strategies offer no clear advantage over linear merging.
While DARE performs comparably, TIES is inconsistent: it works well for Llama-2 on GSM8K, but fails on PubMedQA, Llama-3.1, and both Qwen models (see \refapp{sec:E_4}), making linear merging the most robust strategy overall.
\section{Conclusion}

We proposed SafeMERGE, a lightweight, post-fine-tuning framework for restoring model safety after fine-tuning.
SafeMERGE selectively merges only the degraded layers with corresponding safety-aligned counterparts, rather than the entire model.
Evaluations on four LLMs across several fine-tuning tasks and two independent red-teaming benchmarks show that SafeMERGE consistently outperforms other defenses, achieving optimal trade-offs between task utility and model safety.
Our results highlight selective, layer-wise merging as an effective strategy for restoring safety in fine-tuned LLMs without sacrificing performance.
\section*{Limitations}

\paragraph{Model and Task Selection.}
SafeMERGE is evaluated on four representative LLMs across two downstream tasks in the main body and three additional tasks in appendix.
While this range captures both model diversity and domain shift, extending the evaluation to additional tasks and model families would provide a broader picture of SafeMERGE’s generality.
However, the computational overhead of fine-tuning and safety testing across more tasks and models is non-trivial. 
We also note that our chosen models and datasets from the main body align with those used in prior studies \citep{bhardwaj2024languagemodelshomersimpson,anonymous2024finetuning,hsu2025safelorasilverlining}, enabling fair comparisons despite this limitation.

\paragraph{Safety Evaluations.}
We use Llama-Guard-3-8B \citep{grattafiori2024llama3} to assess safety on two standard red-teaming benchmarks: DirectHarm \citep{lyu2024keeping} and HexPhi \citep{anonymous2024finetuning}.
While this setup enables reproducible and large-scale evaluation, it inherits the limitations of classifier-based safety assessment. 
However, our choice of Llama-Guard-3-8B reflects current best practices in the field, as demonstrated in prior work by \citet{yao2024survey, qi2024safetyalignmentjusttokens, hsu2025safelorasilverlining}.
Nevertheless, to mitigate potential evaluator bias, we cross-validate our results using ShieldGemma-9B \citep{shieldgemma} in \refapp{sec:safety_guard_model}, observing consistent trends with the Llama-Guard model.

\paragraph{Safe Model.}
SafeMERGE relies on a safety-aligned model for layer merging. 
This introduces an additional step that requires fine-tuning on safety data. 
In our experiments we show that this fine-tuning is simple, requires only a small amount of data drawn from publicly available safety datasets (see \refapp{sec:B_2}), and yields a model that is \emph{task-agnostic}. 
As a result, the safe model can be reused across multiple fine-tuning tasks and only needs to be trained once. 
Nevertheless, investigating domain-specific safety datasets for safe model training remains an interesting direction for future work.

\paragraph{Jailbreak Attacks.}
Our work focuses on safety degradation from fine-tuning on benign tasks. 
Our evaluation thus assesses whether models produce harmful outputs when directly prompted with red-teaming instructions, rather than testing their robustness against jailbreak-style attacks \citep{xu2024comprehensivestudyjailbreakattack}. 
We exclude such evaluations for two reasons: (1) our primary goal is to study alignment loss introduced through fine-tuning, not adversarial prompting, and (2) jailbreak evaluations typically require separate attack pipelines, dynamic prompting strategies, and fine-grained response auditing, all of which are beyond the scope of this study.
\section*{Ethics Statement}

While our method helps restore safety through selective merging with a safety-aligned model, it still relies on pre-trained and fine-tuned models that may carry latent biases, unsafe behaviors, or misalignments inherited from the original training data. 
SafeMERGE does not explicitly filter or debias these components.
Thus, further investigation is needed to understand the impact of such inherited biases in the models used during merging.
\section*{Acknowledgments}

This work was supported in part by the German Federal Ministry of Research, Technology and Space (BMFTR) under the national research initiative on 6G communication systems through the research hub 6G-life (Grants 16KISK002 and 16KIS2414), by the Bavarian State Ministry of Science and the Arts through the project Next Generation AI Computing (GAIn), and by the German Research Foundation (DFG) through the Centre for Tactile Internet with Human-in-the-Loop (CeTI) as part of Germany’s Cluster of Excellence Strategy (EXC 2050/2, ID 390696704).
This work was further supported by the LTI-TUM collaboration with MIT, funded by the BMFTR under the 6G-Atlantic Bridges project, and by IBM Research.


\bibliography{bibliography}


\appendix
\section{Related Work}
\label{sec:A_related_work}

Recent literature features numerous defenses to preserve or restore safety alignment in fine-tuned LLMs. 
We refer to \citet{yao2024survey} for a comprehensive survey, while here we discuss representative methods along three stages of intervention.

\paragraph{Alignment-Stage Defenses.}
These solutions aim to make the base model maximally resilient \emph{before} any user-led fine-tuning. 
Techniques include large-scale data filtering and alignment procedures, such as RLHF \citep{ouyang2022training}, to prevent harmful adaptation. 
Representative defenses are Vaccine \citep{huang2024vaccineperturbationawarealignmentlarge}, RepNoise \citep{rosati2024representationnoisingdefencemechanism}, CTRL \citep{liu2024robustifyingsafetyalignedlargelanguage}, TAR \citep{tamirisa2024tamperresistantsafeguardsopenweightllms}, and Booster \citep{huang2024boostertacklingharmfulfinetuning}, which introduce perturbations, adversarial training, or safety constraints to reinforce alignment robustness before fine-tuning.

\paragraph{Fine-Tuning-Stage Defenses.}
These defenses integrate safety alignment measures \emph{during} fine-tuning.
A common approach is to mix safety data into training, as in SafeInstruct \citep{bianchi2024safetytunedllamaslessonsimproving} and VLGuard \citep{zong2024safetyfinetuningalmostcost}, or to apply regularization to safety-anchor model outputs, such as LDIFS \citep{mukhoti2024finetuningcripplefoundationmodel}, Constrained-SFT \citep{qi2024safetyalignmentjusttokens}, and Freeze methods \citep{wei2024assessingbrittlenesssafetyalignment, li2025safety}. 
Additionally, prompt-based safeguards like BEA \citep{wang2024mitigatingfinetuningbasedjailbreak} and PTST \citep{lyu2024keeping} embed safety triggers into prompts to reinforce alignment without modifying model weights. 
Some of these methods require explicit adjustments to the fine-tuning pipeline, often impractical or too complex for black-box fine-tuning with standard open-source libraries.

\paragraph{Post-Fine-Tuning-Stage Defenses.}
Post-training solutions realign a model \emph{after} it has been (potentially unsafely) fine-tuned. 
This is appealing in scenarios where controlling or monitoring the fine-tuning process is infeasible. 
Notable examples include SafeLoRA~\citep{hsu2025safelorasilverlining}, which projects LoRA updates onto a safety subspace derived from a pre-aligned reference model, and RESTA~\citep{bhardwaj2024languagemodelshomersimpson}, which negatively merges a harmful task vector into a compromised model to restore safe behaviors. 
Other methods include SOMF~\citep{yi2024safetyrealignmentframeworksubspaceoriented}, which utilizes masking techniques to realign a fine-tuned model via task vectors, Antidote~\citep{huang2024antidotepostfinetuningsafetyalignment}, which zeroes out harmful weight coordinates to remove undesired responses, and LSSF~\citep{zhou2026lssfsafetyalignmentlarge}, which restores safety via SVD-based low-rank safety subspace fusion followed by linear arithmetic. 
These techniques are particularly useful for fine-tuning-as-a-service scenarios, as they can be applied post-hoc with minimal compute cost.

\ 

Our method, \textbf{SafeMERGE}, fits into the post-training paradigm, specifically drawing from SafeLoRA and RESTA, but taking a more selective, layer-wise approach. 
Instead of globally projecting or adding a single safety vector, SafeMERGE fuses only those LoRA layers whose updates deviate significantly from safety, measured by a cosine similarity criterion. 
By preserving benign layers intact, it achieves a better trade-off between retaining fine-tuned capabilities and restoring safety.
SafeMERGE's selective merging approach is thus principled and motivated by the shortcomings of other post-fine-tuning defenses that either under-restore safety (SafeLoRA) or cause utility loss (RESTA).
Moreover, SafeMERGE is intentionally simple, requires no retraining, and is easy to tune.
This distinguishes it from more resource-intensive post-fine-tuning-stage approaches such as LSSF, which relies on matrix decomposition (SVD) to identify low-rank safety subspaces, whereas SafeMERGE operates through cosine-similarity checks and linear blending, making it more accessible and mechanistically interpretable.
\section{LoRA Fine-Tuning}
\label{sec:lora}

Low-Rank Adaptation (LoRA)~\cite{hu2021loralowrankadaptationlarge} is a parameter-efficient fine-tuning (PEFT) method that enables large pre-trained language models to be fine-tuned efficiently with minimal additional parameters.
Instead of updating the full parameter set of a model during fine-tuning, LoRA injects a pair of small, trainable rank-decomposition matrices into selected transformer layers while keeping the original weights frozen.
This approach drastically reduces the number of trainable parameters and corresponding memory footprint, while maintaining competitive performance to full fine-tuning.
Its simplicity and compute efficiency make LoRA one of the most widely adopted techniques for parameter-efficient fine-tuning of LLMs.

\subsection{General Formulation}
Consider a weight matrix $W^i \in \mathbb{R}^{d \times k}$ of a transformer layer $i$, e.g., one of the projection matrices in a self-attention or feed-forward block.
In standard fine-tuning, all entries of $W^i$ are updated.
In contrast, LoRA constrains the weight update $\Delta W^i$ to be a low-rank decomposition, i.e.,
\begin{equation}
    \Delta W^i = A^i B^i,
\end{equation}
where $A^i \in \mathbb{R}^{d \times r}$ and $B^i \in \mathbb{R}^{r \times k}$ are trainable matrices of rank $r$ with $r \ll \min(d, k)$.
The adapted LoRA weight matrix is thus given by
\begin{equation}
    W^i_{\text{LoRA}} = W^i + \gamma \cdot A^i B^i ,
\end{equation}
where $\gamma$ is a scalar scaling factor introduced to stabilize training by controlling the effective magnitude of the weight update.

\subsection{Forward and Backward Computation}
During forward propagation, the modified linear transformation becomes
\begin{equation}
    h_{\text{out}} = W^i_{\text{LoRA}} h_{\text{in}} = W^i h_{\text{in}} + \gamma \cdot A^i (B^i h_{\text{in}}),
\end{equation}
where $h_{\text{in}}$ and $h_{\text{out}}$ denote the input and output activations, respectively.
Since $W^i$ remains frozen, gradients are only computed with respect to $A^i$ and $B^i$ during backpropagation, i.e.,
\begin{align}
    \frac{\partial \mathcal{L}}{\partial A^i} &= \gamma \cdot \frac{\partial \mathcal{L}}{\partial h_{\text{out}}} (B^i h_{\text{in}})^\top, \\
    \frac{\partial \mathcal{L}}{\partial B^i} &= \gamma \cdot (A^i)^\top \frac{\partial \mathcal{L}}{\partial h_{\text{out}}} h_{\text{in}}^\top,
\end{align}
where $\mathcal{L}$ denotes the training loss function.

\subsection{Parameter Efficiency}
The total number of trainable parameters for an adapted LoRA matrix is given by
\begin{equation}
    N_{\text{LoRA}} = dr + rk = r(d + k),
\end{equation}
compared to $dk$ for full fine-tuning.
For typical configurations where $r \in \{4, 8, 16\}$, LoRA achieves $\approx$0.1--1\% trainable parameters relative to the base model, depending on which layers are adapted.

\subsection{Application to Linear Projections}
LoRA is typically applied to selected linear projections in transformer layers, such as the attention projection matrices $\{W_q, W_k, W_v, W_o\}$ or the feed-forward (MLP) projections.
Applying LoRA to a subset of these modules, most commonly $W_q$ and $W_v$, offers a favorable trade-off between model quality and training efficiency.
In practice, applying LoRA across all attention and feed-forward modules yields only marginal performance gains while increasing training cost.
We refer readers to the original work by \citet{hu2021loralowrankadaptationlarge} for further theoretical and empirical details.
\section{Fine-Tuning Configurations}
\label{sec:B_fine_tuning}

This section provides details on the fine-tuning configurations used in our experiments.
To ensure reproducibility and comparability, we fine-tune all models using Llama-Factory \citep{zheng2024llamafactory} with FSDP on 8 $\times$ NVIDIA A100 80GB GPUs.

\subsection{Utility Fine-Tuning}
\label{sec:utility_fine_tuning}
We LoRA fine-tune \citep{hu2021loralowrankadaptationlarge} Llama-2-7B-Chat, Llama-3.1-8B-Instruct, Qwen-2-7B-Instruct, and Qwen-2.5-7B-Instruct on GSM8K and PubMedQA tasks using the configurations detailed in Table \ref{tab:lorasettings}.
This results in $\approx1$\% trainable parameters, making fine-tuning more efficient without compromising accuracy.
Fine-tuning details for the additional telecom datasets are provided in \refapp{sec:telecom_results}.

\begin{table}[htbp]
\centering
\caption{Fine-tuning hyperparameters for GSM8K and PubMedQA across Llama-2, Llama-3.1, Qwen-2, and Qwen-2.5 models.}
\label{tab:lorasettings}
\resizebox{0.4\textwidth}{!}{
    \begin{tabular}{lcc}
        \toprule
        \textbf{Parameter} & \textbf{GSM8K} & \textbf{PubMedQA} \\
        \midrule
        Batch Size         & 32             & 64 \\
        Learning Rate      & $1\times10^{-4}$ & $1\times10^{-4}$ \\
        Epochs             & 6              & 2 \\
        Warmup             & 64 steps       & 1\% of total steps \\
        LR Scheduler       & Linear         & Cosine \\
        Weight Decay       & 0              & 0.01 \\
        \midrule
        LoRA Modules     & [q\_proj, v\_proj] & [q\_proj, v\_proj] \\
        LoRA Rank               & 8                & 8 \\
        LoRA Alpha              & 16               & 16 \\
        LoRA Dropout            & 0                & 0 \\
        \bottomrule
    \end{tabular}
}
\end{table}

\subsection{Safety Fine-Tuning}
\label{sec:B_2}
We similarly fine-tune the corresponding safety models on 100, 500, 1000, and 2500 samples from \citet{bianchi2024safetytunedllamaslessonsimproving}'s safety collection using the LoRA parameters from Table \ref{tab:lorasettings} with batch size 32, learning rate $1 \times 10^{-4}$, and linear scheduling for 10 epochs each. 
We then select the best (i.e., the safest) model.
Tables \ref{tab:safety_model_harmfulness_scores_llama} and \ref{tab:safety_model_harmfulness_scores_qwen} report the harmfulness scores after safety fine-tuning for all models.

\begin{table}[htbp]
    \centering
    \caption{Harmfulness scores (lower is better) for safe Llama-2 and Llama-3.1 models across different safety sample sizes from \citet{bianchi2024safetytunedllamaslessonsimproving}.}
    \label{tab:safety_model_harmfulness_scores_llama}
    \resizebox{0.4\textwidth}{!}{
        \begin{tabular}{lcccc}
            \toprule
             & \multicolumn{2}{c}{\textbf{Llama-2-7B-Chat}} & \multicolumn{2}{c}{\textbf{Llama-3.1-8B-Instruct}} \\
            \cmidrule(lr){2-3} \cmidrule(lr){4-5}
            \textbf{Samples} & DirectHarm & HexPhi & DirectHarm & HexPhi \\
            \midrule
            100  & 3.00 & 2.30 & 8.90 & 6.10 \\
            500  & 1.80 & 2.60 & 7.10 & 5.70 \\
            1000 & \textbf{1.30} & \textbf{1.00} & \textbf{6.30} & \textbf{5.10} \\
            2500 & 1.50 & 2.00 & 6.40 & 5.20 \\
            \bottomrule
        \end{tabular}
    }
\end{table}

\begin{table}[htbp]
    \centering
    \caption{Harmfulness scores (lower is better) for safe Qwen-2 and Qwen-2.5 models across different safety sample sizes from \citet{bianchi2024safetytunedllamaslessonsimproving}.}
    \label{tab:safety_model_harmfulness_scores_qwen}
    \resizebox{0.4\textwidth}{!}{
        \begin{tabular}{lcccc}
            \toprule
             & \multicolumn{2}{c}{\textbf{Qwen-2-7B-Instruct}} & \multicolumn{2}{c}{\textbf{Qwen-2.5-7B-Instruct}} \\
            \cmidrule(lr){2-3} \cmidrule(lr){4-5}
            \textbf{Samples} & DirectHarm & HexPhi & DirectHarm & HexPhi \\
            \midrule
            100  & 15.50 & 9.90 & 11.10 & 10.90 \\
            500  & \textbf{6.80} & \textbf{3.30} & \textbf{5.80} & \textbf{3.00} \\
            1000 & 7.50 & 3.00 & 6.30 & 3.10 \\
            2500 & 9.20 & 6.90 & 8.70 & 5.40 \\
            \bottomrule
        \end{tabular}
    }
\end{table}
\section{Evaluation Setup}
\label{sec:C_evaluation_setup}

This section details the utility and safety evaluation setup used in our experiments.

\subsection{Utility Evaluations}
\label{sec:C_1}
To assess task performance on the \emph{primary utility datasets}, we use the \emph{LM Evaluation Harness} framework \citep{lm-eval-harness}, a widely adopted standard for evaluating LLMs across diverse benchmarks.
To this end, we report exact-match accuracy for GSM8K (0-shot) and classification accuracy for PubMedQA.
To further assess general reasoning abilities, we evaluate each model on IFEval \citep{zhou2023instructionfollowingevaluationlargelanguage} and MMLU \citep{hendryckstest2021}.
For all evaluations, we follow the default configuration and settings from the LM Evaluation Harness framework.
Evaluation details for the additional telecom datasets are provided in \refapp{sec:telecom_results}.

\subsection{Safety Evaluations}
\label{sec:C_2}
For each model, we generate responses to harmful prompts from the DirectHarm \citep{lyu2024keeping} and HexPhi \citep{anonymous2024finetuning} red-teaming datasets, and evaluate their corresponding safety using Llama-Guard-3-8B \citep{grattafiori2024llama3}.
To this end, we report the overall harmfulness score as the proportion of model responses flagged as unsafe.
Table~\ref{tab:inference_params} summarizes the inference parameters used for response generation across all models.
To avoid potential evaluator bias, we cross-validate our safety scores using ShieldGemma-9B \citep{shieldgemma}, observing consistent trends with our Llama guard model (see \refapp{sec:safety_guard_model}).

\begin{table}[htbp]
    \centering
    \caption{Inference parameters used for generating responses to harmful prompts across all models.}
    \label{tab:inference_params}
    \resizebox{0.25\textwidth}{!}{
    \begin{tabular}{lc}
        \toprule
        \textbf{Parameter}       & \textbf{Value} \\
        \midrule
        max\_new\_tokens         & 512 \\
        top\_p                   & 1.0 \\
        top\_k                   & 0 \\
        temperature              & 1.0 \\
        repetition\_penalty      & 1.0 \\
        length\_penalty          & 1 \\
        batch\_size              & 1 \\
        \bottomrule
    \end{tabular}
    }
\end{table}
\section{Baseline Configurations and Results}
\label{sec:D_baselines}

This section provides details on the selected baseline defenses, their optimal configurations, and additional intermediate results.

\subsection{SafeInstruct}
\label{sec:D_safeinstruct}
Following \citet{bianchi2024safetytunedllamaslessonsimproving}, we randomly interleave a set of their \href{https://github.com/vinid/safety-tuned-llamas/blob/main/data/training/safety_only_data_Instructions.json}{harmful Q\&A pairs (with safe answers)} into the corresponding fine-tuning datasets without additional system prompts. 
We experiment with 100, 500, 1000, and 2500 interleaved safety samples. Since the total number of safety samples remains relatively small (e.g., at most 1.2\% of PubMedQA and 28\% of GSM8K), we retain the original downstream task fine-tuning hyperparameters from Table~\ref{tab:lorasettings}.

\subsubsection{Fine-Tuning Results}
\label{sec:D_1}
In general, we confirm \citet{bianchi2024safetytunedllamaslessonsimproving}'s observation that more samples increase safety, and even may increase utility.
We report intermediate results for Llama-2 and Llama-3.1 models in Table~\ref{tab:safeinstruct_comparison_llama}, and for Qwen-2 and Qwen-2.5 models in Table~\ref{tab:safeinstruct_comparison_qwen}. 
For comparison with SafeMERGE, we select the safest variant, i.e., the one with 2500 safety samples.

\begin{table*}[htbp]
\centering
\caption{SafeInstruct at various safety sample sizes for Llama-2-7B-Chat and Llama-3.1-8B-Instruct.}
\label{tab:safeinstruct_comparison_llama}
\resizebox{\textwidth}{!}{%
    \begin{tabular}{ccccccccccccc}
    \toprule
    & \multicolumn{6}{c}{\textbf{Llama-2-7B-Chat}} 
    & \multicolumn{6}{c}{\textbf{Llama-3.1-8B-Instruct}} \\
    \cmidrule(lr){2-7}\cmidrule(lr){8-13}
    & \multicolumn{3}{c}{GSM8K} 
    & \multicolumn{3}{c}{PubMedQA} 
    & \multicolumn{3}{c}{GSM8K}
    & \multicolumn{3}{c}{PubMedQA} \\
    \cmidrule(lr){2-4}\cmidrule(lr){5-7}\cmidrule(lr){8-10}\cmidrule(lr){11-13}
    \makecell[c]{\textbf{SafeInstruct} \\ Number of Samples}
    & DirectHarm & HexPhi & Utility
    & DirectHarm & HexPhi & Utility
    & DirectHarm & HexPhi & Utility
    & DirectHarm & HexPhi & Utility \\
    \midrule
    100   & 10.20 & 10.50 & 23.42 & 26.00 & 10.90 & 69.40 & 17.80 & 15.70 & 76.90 & 18.90 & 12.20 & 78.30 \\
    500   & 10.00 & 7.90 & 23.80 & 18.80 & 10.50 & 69.70 & 14.40 & 9.40 & 76.80 & 15.10 & 11.20 & 77.90 \\
    1000  & 7.90 & 6.80 & 25.17 & 15.20 & 6.90 & 71.20 & 13.70 & 7.80 & 77.20 & 12.50 & 10.40 & 78.10 \\
    2500  & \textbf{7.50} & \textbf{6.20} & \textbf{26.00} & \textbf{12.20} & \textbf{6.30} & \textbf{71.20} & \textbf{12.50} & \textbf{7.20} & \textbf{77.40} & \textbf{11.80} & \textbf{9.70} & \textbf{78.50} \\
    \bottomrule
    \end{tabular}
}
\end{table*}

\begin{table*}[htbp]
\centering
\caption{SafeInstruct at various safety sample sizes for Qwen-2-7B-Instruct and Qwen-2.5-7B-Instruct.}
\label{tab:safeinstruct_comparison_qwen}
\resizebox{\textwidth}{!}{%
    \begin{tabular}{ccccccccccccc}
    \toprule
    & \multicolumn{6}{c}{\textbf{Qwen-2-7B-Instruct}} 
    & \multicolumn{6}{c}{\textbf{Qwen-2.5-7B-Instruct}} \\
    \cmidrule(lr){2-7}\cmidrule(lr){8-13}
    & \multicolumn{3}{c}{GSM8K} 
    & \multicolumn{3}{c}{PubMedQA} 
    & \multicolumn{3}{c}{GSM8K}
    & \multicolumn{3}{c}{PubMedQA} \\
    \cmidrule(lr){2-4}\cmidrule(lr){5-7}\cmidrule(lr){8-10}\cmidrule(lr){11-13}
    \makecell[c]{\textbf{SafeInstruct} \\ Number of Samples}
    & DirectHarm & HexPhi & Utility
    & DirectHarm & HexPhi & Utility
    & DirectHarm & HexPhi & Utility
    & DirectHarm & HexPhi & Utility \\
    \midrule
    100   & 19.50 & 10.90 & 71.42 & 15.50 & 6.30 & 79.20 & 17.40 & 15.60 & 77.10 & 16.10 & 14.40 & 78.60 \\
    500   & 17.50 & 10.20 & 72.07 & 14.20 & 5.90 & 79.60 & 15.50 & 15.20 & 77.20 & 14.90 & 12.20 & 78.20 \\
    1000  & 15.70 & 9.90  & 72.42 & 13.50 & 5.90 & 79.20 & 13.30 & 12.50 & 77.60 & 13.80 & 10.30 & 78.90 \\
    2500  & \textbf{13.70} & \textbf{9.50} & \textbf{72.69} & \textbf{12.50} & \textbf{5.90} & \textbf{80.00} 
          & \textbf{12.50} & \textbf{11.30} & \textbf{77.80} & \textbf{13.10} & \textbf{9.50} & \textbf{79.20} \\
    \bottomrule
    \end{tabular}
}
\end{table*}

\subsubsection{Utility vs. Safety Trade-Off}
\begin{figure}[t]
  \centering
  \includegraphics[width=0.48\textwidth]{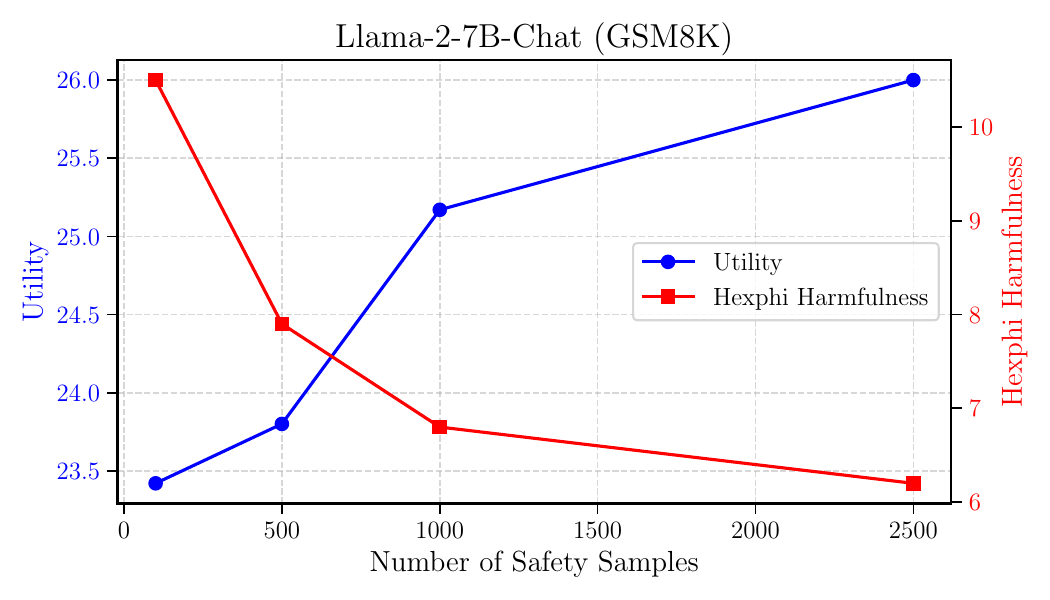}
  \caption{SafeInstruct utility vs. safety for Llama-2-7B-Chat (GSM8K), evaluated on HexPhi prompts.}
  \label{fig:llama_2_gsm8k_hexphi}
\end{figure}

\begin{figure}[t]
  \centering
  \includegraphics[width=0.48\textwidth]{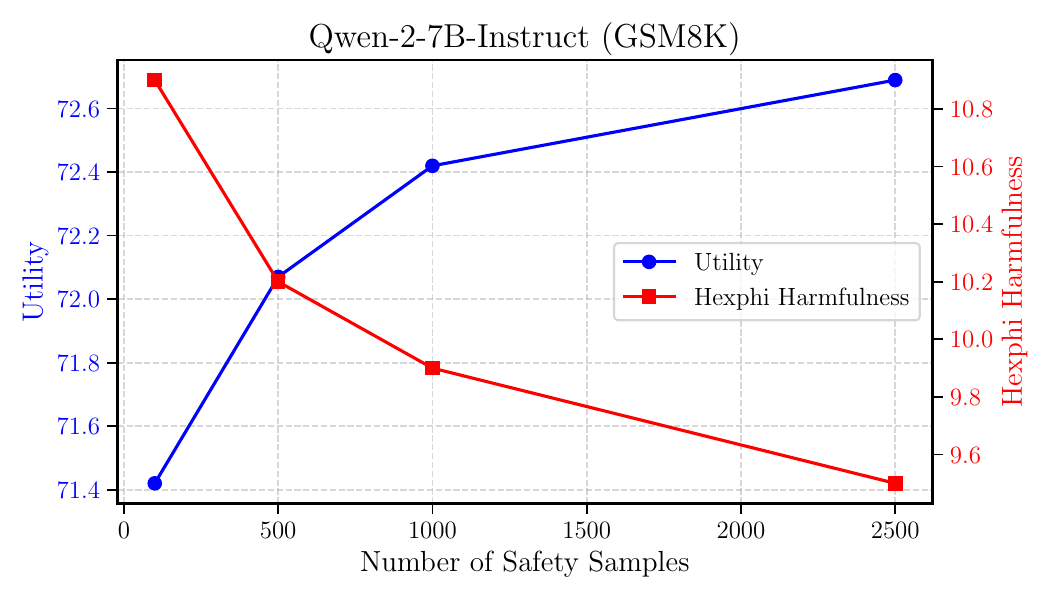}
  \caption{SafeInstruct utility vs. safety for Qwen-2-7B-Instruct (GSM8K), evaluated on HexPhi prompts.}
  \label{fig:qwen_gsm8k_hexphi}
\end{figure}

\refig{fig:llama_2_gsm8k_hexphi}–\ref{fig:qwen_gsm8k_hexphi} compare utility (blue, left $y$-axis) and HexPhi harmfulness (red, right $y$-axis) across different safety sample sizes for GSM8K.
This illustrates the trade-off between utility and harmfulness, corroborating the findings of \citet{bianchi2024safetytunedllamaslessonsimproving}.
The patterns are similar for Llama-3.1 and Qwen-2.5.

\subsection{RESTA}
\label{sec:D_resta}
RESTA~\citep{bhardwaj2024languagemodelshomersimpson} constructs a safety vector by fine-tuning a model on harmful data and negating the resulting LoRA parameters. 
Since the original dataset used in \citet{bhardwaj2024languagemodelshomersimpson} is unavailable, we replicate RESTA using AdvBench \citep{zou2023universal} and HarmfulQA \citep{bhardwaj2023redteaming}. 
We evaluate both linear and DARE-linear merging and explore densities from 0.1 to 0.5, as well as weighting factors \( \alpha \in [0.1, 0.5] \).

\subsubsection{Implementation}
The RESTA methodology follows the below steps:

\begin{enumerate}
    \item Fine-tune a harmful model using Adv-Bench/HarmfulQA, resulting in \( \theta_{\text{harmful}} \).
    
    \item Negate all harmful LoRA weights, resulting in \( \theta^*_{\text{harmful}} \), i.e., perform for all LoRAs:
    \begin{align*}
    W_{\text{harm}}^{\text{LoRA}, *} = -W_{\text{harm}}^{\text{LoRA}} .
    \end{align*}
    
    \item Merge the negated weights \( \theta^*_{\text{harmful}} \) with the original fine-tuned model $\theta_{\text{SFT}}^{orig}$, i.e.,
    \begin{align*}
    \theta_{\text{RESTA}} = \theta_{\text{SFT}}^{orig} + \alpha \cdot \theta^*_{\text{harmful}} ,
    \end{align*}
    where \( \alpha \in [0.1, 0.5] \) is the weighting factor.
    
    \item Apply DARE rescaling if required.
\end{enumerate}

We implement the corresponding LoRA adapter merging using \href{https://huggingface.co/docs/peft/en/index}{HuggingFace's PEFT library}, which supports both linear and DARE-linear merging.

\subsubsection{Harmful Fine-Tuning}
We fine-tune Llama and Qwen models on Adv-Bench and HarmfulQA datasets using the LoRA settings from Table~\ref{tab:lorasettings} with a batch size of 32, learning rate of $1 \times 10^{-4}$, and linear scheduling for 5 epochs. 
We report the harmfulness scores for DirectHarm and HexPhi in Table~\ref{tab:resta_harmfulness_scores}.

\begin{table}[htbp]
    \centering
    \caption{Harmfulness scores (higher is better for RESTA) for Llama-2, Llama-3.1, Qwen-2, and Qwen-2.5 models across DirectHarm and HexPhi benchmarks.}
    \label{tab:resta_harmfulness_scores}
    \resizebox{0.48\textwidth}{!}{
    \begin{tabular}{lcccc}
        \toprule
         & \multicolumn{2}{c}{\textbf{AdvBench}} & \multicolumn{2}{c}{\textbf{HarmfulQA}} \\
        \cmidrule(lr){2-3} \cmidrule(lr){4-5}
        \textbf{Model} & DirectHarm & HexPhi & DirectHarm & HexPhi \\
        \midrule
        Llama-2-7B-Chat         & 38.30 & 36.50 & 94.00 & 97.40 \\
        Llama-3.1-8B-Instruct   & 64.50 & 62.70 & 95.50 & 98.20 \\
        Qwen-2-7B-Instruct      & 59.50 & 47.70 & 72.00 & 76.00 \\
        Qwen-2.5-7B-Instruct    & 64.80 & 53.60 & 77.10 & 78.30 \\
        \bottomrule
    \end{tabular}
    }
\end{table}

\begin{figure*}[t]
  \centering
  \begin{subfigure}[b]{0.48\linewidth}
    \centering
    \includegraphics[width=\linewidth]{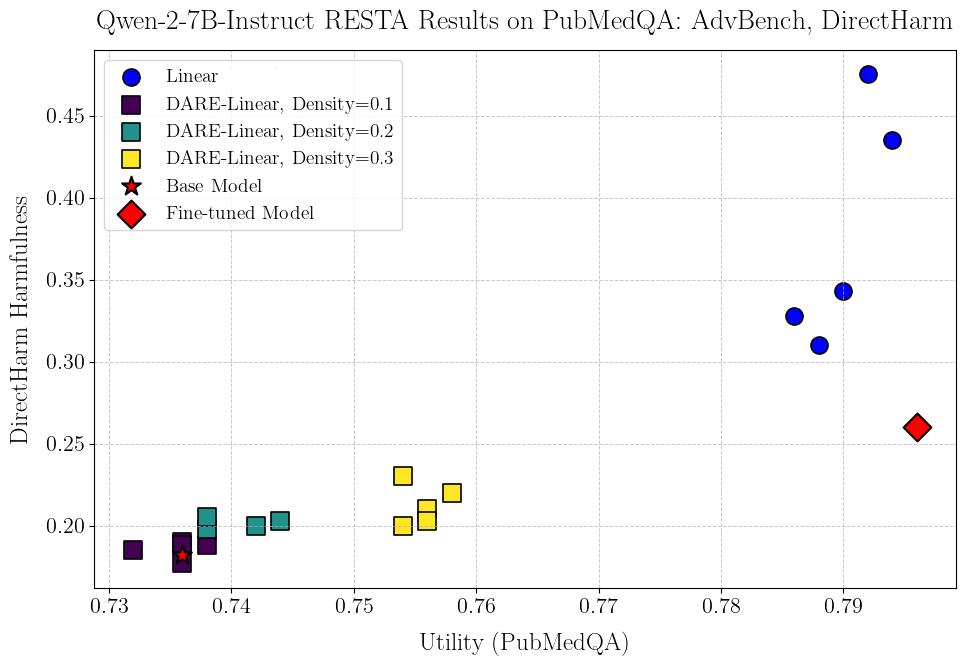}
    \caption{AdvBench: Utility vs. DirectHarm Harmfulness.}
    \label{fig:advbench_directharm}
  \end{subfigure}
  \hfill
  \begin{subfigure}[b]{0.48\linewidth}
    \centering
    \includegraphics[width=\linewidth]{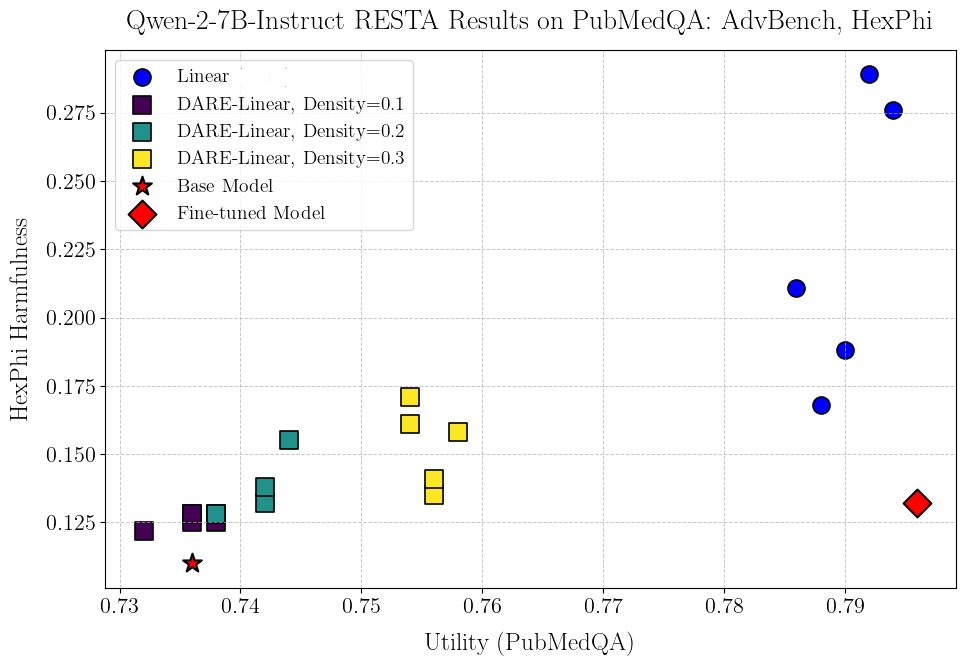}
    \caption{AdvBench: Utility vs. HexPhi Harmfulness.}
    \label{fig:advbench_hexphi}
  \end{subfigure}
  
  \vspace{1em}

  \begin{subfigure}[b]{0.48\linewidth}
    \centering
    \includegraphics[width=\linewidth]{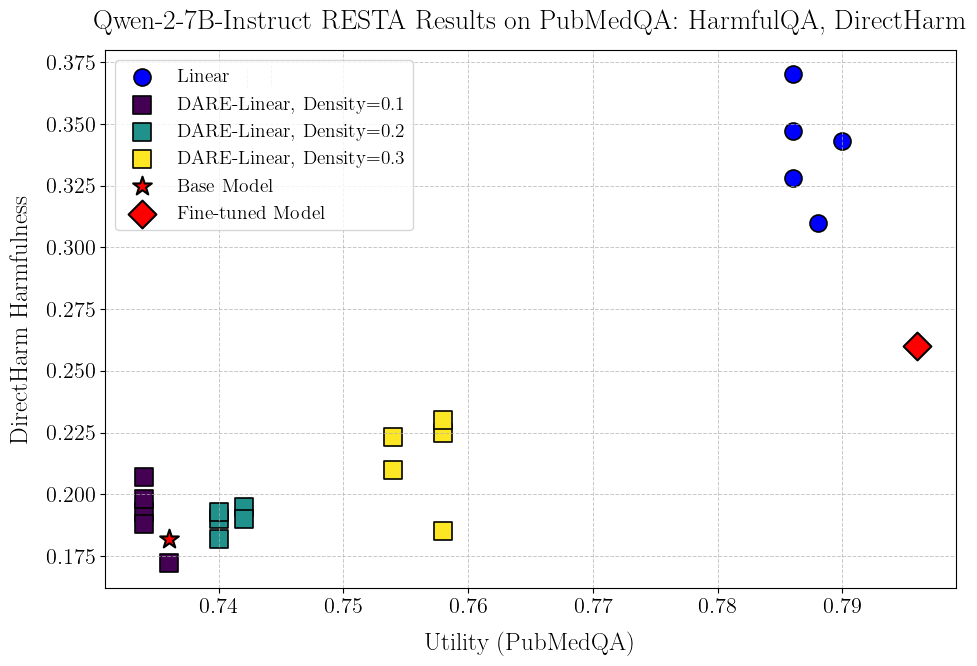}
    \caption{HarmfulQA: Utility vs. DirectHarm Harmfulness.}
    \label{fig:harmfulqa_directharm}
  \end{subfigure}
  \hfill
  \begin{subfigure}[b]{0.48\linewidth}
    \centering
    \includegraphics[width=\linewidth]{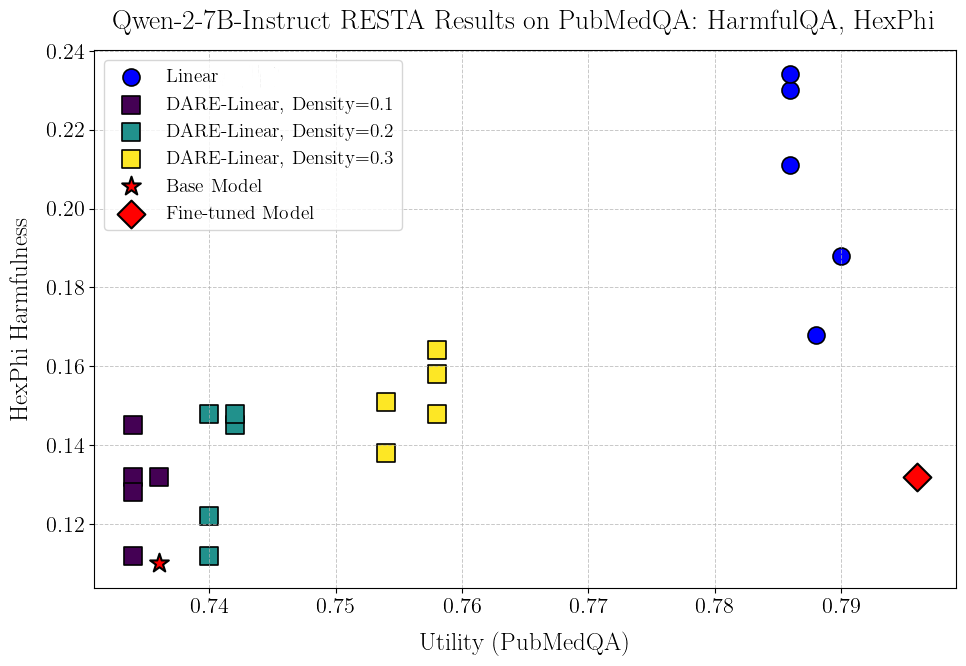}
    \caption{HarmfulQA: Utility vs. HexPhi Harmfulness.}
    \label{fig:harmfulqa_hexphi}
  \end{subfigure}

  \caption{Utility vs. safety trade-offs of RESTA for Qwen-2-7B-Instruct across experiments. The first row shows results on AdvBench, the second row on HarmfulQA. Each point corresponds to a different weighting \( \alpha \in [0.1, 0.5] \).}

  \label{fig:qwen_2x2_results}
\end{figure*}

\subsubsection{RESTA Weighting Factors vs. Density vs. Linear vs. DARE-linear Merging}
\label{sec:D_2.3}
We analyze the trade-off between utility and safety for different weightings under both linear and DARE-linear merging in RESTA.
Figures~\ref{fig:advbench_directharm}--\ref{fig:harmfulqa_hexphi} present results for Qwen-2-7B-Instruct fine-tuned on AdvBench and HarmfulQA, evaluated on PubMedQA for utility and on DirectHarm and HexPhi for harmfulness. 
We find that DARE-linear merging consistently yields better safety scores but at the expense of lower task performance.
In general, lower densities further worsen this trade-off.  
By contrast, linear merging remains close in utility to the fine-tuned model but compromises safety, significantly increasing harmfulness.  
Each point in the figures corresponds to a different weighting factor, where we identify \( \alpha = 0.5 \) as the best among them. 
Similar patterns are observed for Llama-2, Llama-3.1, and Qwen-2.5 models but are omitted for brevity.  
Overall, RESTA’s approach to realigning safety compromises either task utility or safety, suggesting that attempting to remove harmful task vectors through negative merging is less effective, at least for the evaluated utility tasks in our study.

\subsection{RESTA-Instruct}
\label{sec:D_resta_instruct}
Similar to RESTA, \citet{farn2025safeguardfinetunedllmspre} perform full-parameter merging between fine-tuned and instruct models, aiming to induce safe task vectors instead of attempting to remove harmful ones, i.e.,
\begin{align*}
    \theta_{\text{merged}} = \theta_{\text{SFT}}^{orig} + \alpha \cdot \theta_{\text{instruct}} .
\end{align*}
We refer to this variant as \emph{RESTA-Instruct} and conduct similar experiments, where we identify \( \alpha = 0.5 \) as the best weighting factor. 
In general, we observe similar general trends compared to RESTA, with RESTA-Instruct, on average, achieving a slightly better utility.
However, as shown in Table~\ref{tab:consolidated_results}, RESTA-Instruct continues to compromise on task utility compared to other defenses, further suggesting that merging with the instruct model erases task vectors learned during fine-tuning, eventually inducing catastrophic forgetting.
Furthermore, we emphasize that this special case of RESTA assumes that instruct models are inherently safety-aligned, which is not the case for all models, such as Mistral-7B-Instruct \citep{jiang2023mistral7b}.
This may create a serious oversight.
Adding to that, we observe that, for example, the Llama-3.1-8B-Instruct model is noticeably less safe compared to Llama-2-7B-Chat. 
Thus, the incorporation of a safety-aligned model, as is done for SafeMERGE, is a better choice to effectively restore safety.
In addition, naively merging all parameters may be too simplistic, further encouraging catastrophic forgetting.

\subsection{SafeLoRA}
\label{sec:D_safelora}
SafeLoRA~\citep{hsu2025safelorasilverlining} mitigates safety degradation in fine-tuned models by projecting LoRA weight updates onto a safety-aligned subspace. 
We apply SafeLoRA to Llama-2, Llama-3.1, Qwen-2, and Qwen-2.5 models, using their respective base and instruct variants to construct the safety-aligned subspace.  
To this end, we tune the cosine similarity threshold \( \tau \) between 0.1 and 1.0.

\subsubsection{Implementation}
\label{sec:safelora_implementation}
We follow the implementation provided in the official repository of \citet{hsu2025safelorasilverlining}. 
The per-layer projection matrix \( C^i \) is computed using the respective instruct/chat and base variants of each model.
In general, we find that instruct/chat models are already well safety-aligned, sufficiently capturing harmful directions when comparing layers via cosine similarity.  
Nevertheless, for Qwen-2, we investigate two approaches to construct the safety-aligned subspace: (i) using the base model, and (ii) using a safety-tuned model with 500 safety samples from \citet{bianchi2024safetytunedllamaslessonsimproving}.
Results show that most LoRA projections remain identical across both approaches, suggesting that differences are primarily reflected in the scaling of \( \tau \).  
In contrast, using an off-the-shelf instruct model is not sufficient.
This observation directly transfers to SafeMERGE.

\subsubsection{Threshold Selection and Projected LoRA Layers}
\label{sec:D_3.2}
\begin{figure}[!t]
  \centering
  \includegraphics[width=0.45\textwidth]{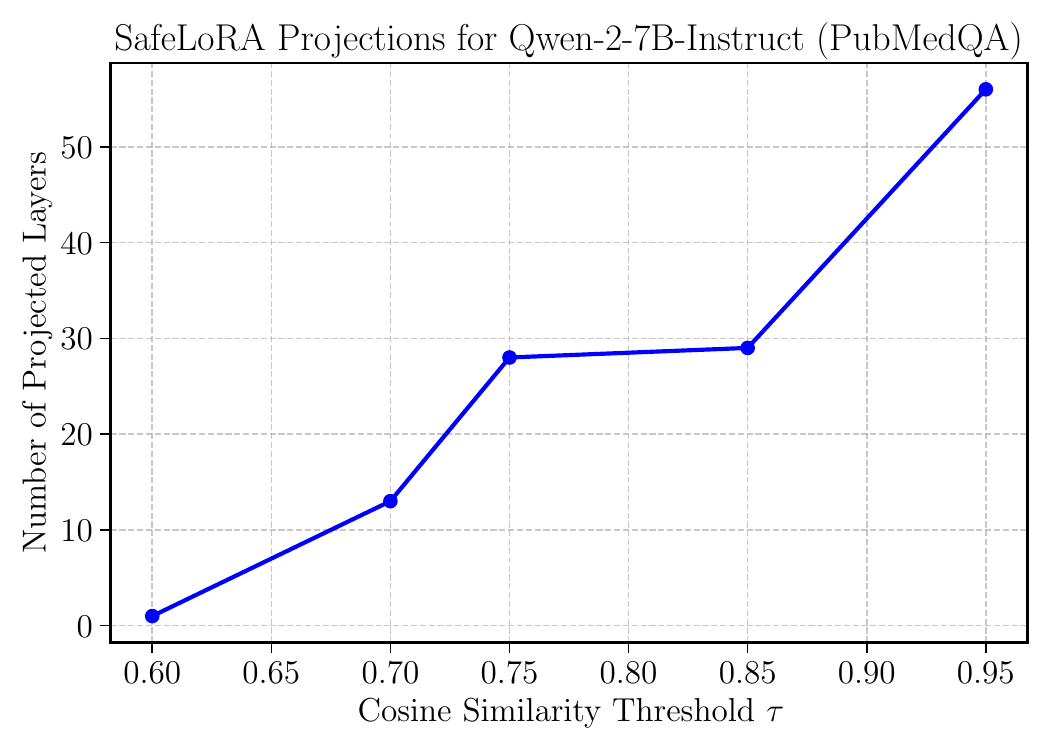}
  \caption{SafeLoRA projections for Qwen-2-7B-Instruct (PubMedQA) as a function of threshold \( \tau \).}
  \label{fig:safelora_qwen_projections}
\end{figure}

We analyze the threshold factor \( \tau \) and the number of projected layers in SafeLoRA.
Due to the LoRA formulation, projection is required for only one of the two trainable LoRA components (\emph{LoRA-A} or \emph{LoRA-B}) since multiplication with the other inherently incorporates the projection.  
Formally, given the LoRA update \( \Delta W^i = A^i B^i \),
SafeLoRA applies the projection only to one component, e.g.,
\begin{align*}
    \Delta W_{\text{proj}}^i = (C^i A^i) B^i \quad \text{or} \quad \Delta W_{\text{proj}}^i = A^i (B^i C^i),
\end{align*}
where \( C^i \) denotes the projection matrix for layer \( i \).
Thus, the maximum number of projected layers is 56 for Qwen-2/2.5 models and 64 for Llama-2 and Llama-3.1 models.  
\refig{fig:safelora_qwen_projections} illustrates how layers are progressively projected as the threshold \( \tau \) increases for Qwen-2-7B-Instruct fine-tuned on PubMedQA.
In general, lower thresholds result in fewer projected layers, preserving downstream task performance but limiting safety improvements.

\subsubsection{Projection vs. Harmfulness vs. Utility}
\label{sec:D_3.3}
\begin{figure}[!t]
    \centering
    \includegraphics[width=0.48\textwidth]{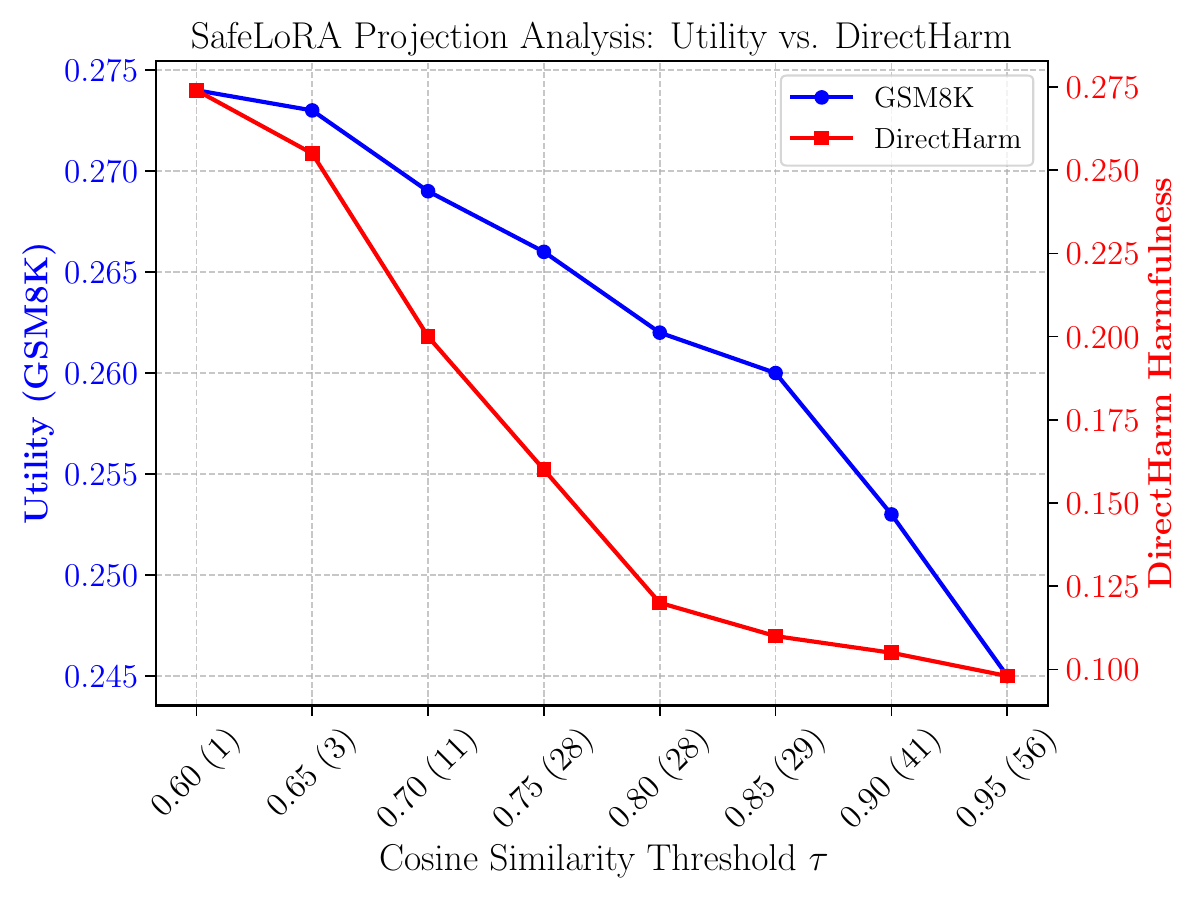}
    \caption{SafeLoRA projections vs. harmfulness (DirectHarm) vs. utility (GSM8K) for Llama-2-7B-Chat.}
    \label{fig:safelora_llama_gsm8k_plot}
    \vspace{-0.5cm}
\end{figure}

We compare SafeLoRA's performance with the number of projected layers and harmfulness (DirectHarm) for Llama-2-7B-Chat (GSM8K) in \refig{fig:safelora_llama_gsm8k_plot}.
In general, as more layers are projected, task utility decreases while safety improves.
Selecting a balanced cosine similarity threshold \( \tau \) is therefore

\noindent
important to optimize the trade-off between task utility and safety.
Similar trends are observed for PubMedQA as well as for the Llama-3.1, Qwen-2, and Qwen-2.5 models.
SafeLoRA, on average, retains higher utility on challenging datasets but reduces harmfulness less effectively than SafeInstruct or RESTA variants. 
This motivates SafeMERGE to employ layer-wise merging to achieve a more effective balance between safety and performance.
\begin{figure}[H]
  \centering

  \begin{subfigure}[t]{0.99\linewidth}
    \centering
    \includegraphics[width=\linewidth]{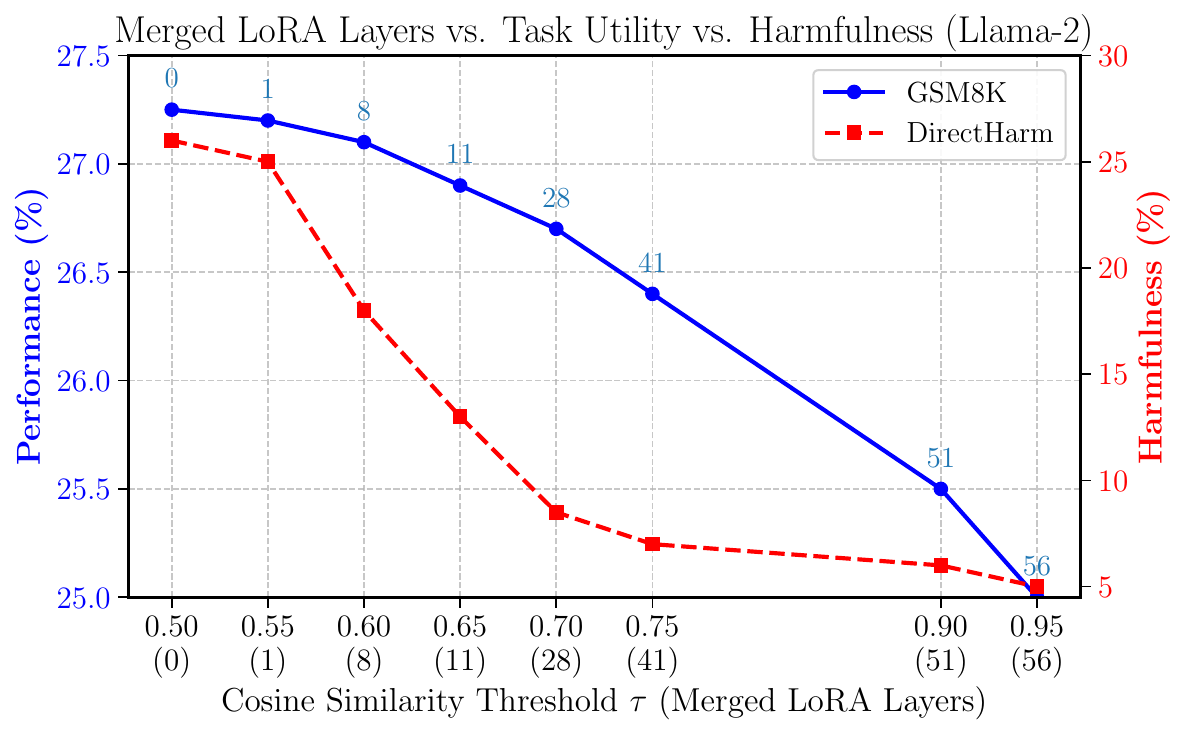}
    \caption{Llama-2-7B-Chat (GSM8K, DirectHarm) with weighting factors $[0.8, 0.2]$, i.e., \( \alpha = 0.8 \). The optimal trade-off between utility and safety is achieved for \( \tau = 0.7 \).}
    \label{fig:safemerge_threshold_analysis_llama2_gsm8k}
  \end{subfigure}

  \vspace{0.5em}

  \begin{subfigure}[t]{0.99\linewidth}
    \centering
    \includegraphics[width=\linewidth]{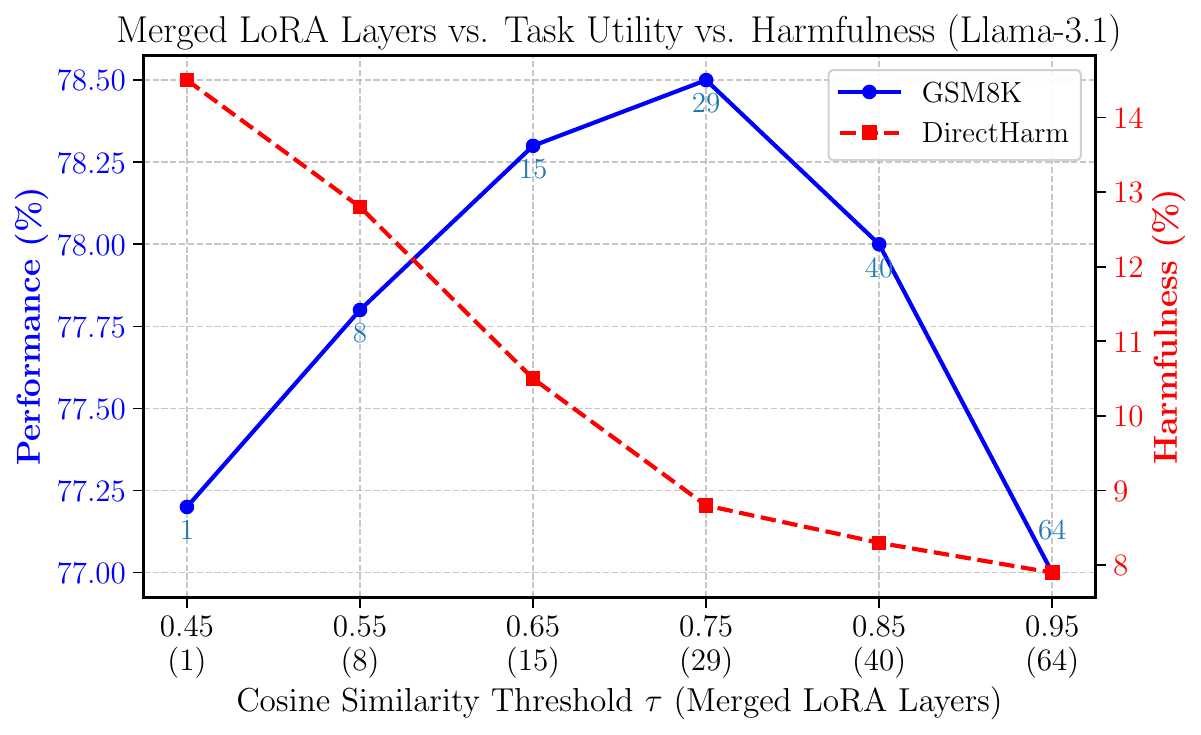}
    \caption{Llama-3.1-8B-Instruct (GSM8K, DirectHarm) with weighting factors $[0.8, 0.2]$, i.e., \( \alpha = 0.8 \). The optimal trade-off between utility and safety is achieved for \( \tau = 0.75 \).}
    \label{fig:safemerge_threshold_analysis_llama3_gsm8k}
  \end{subfigure}

  \vspace{0.5em}

  \begin{subfigure}[t]{0.99\linewidth}
    \centering
    \includegraphics[width=\linewidth]{figures/safemerge_threshold_analysis_qwen2_gsm8k_directharm.pdf}
    \caption{Qwen-2-7B-Instruct (GSM8K, DirectHarm) with weighting factors $[0.7, 0.3]$, i.e., \( \alpha = 0.7 \). The optimal trade-off between utility and safety is achieved for \( \tau = 0.65 \).}
    \label{fig:safemerge_threshold_analysis_qwen2_gsm8k}
  \end{subfigure}

  \caption{SafeMERGE performance for different thresholds \( \tau \) on Llama-2, Llama-3.1, and Qwen-2 models fine-tuned on GSM8K. As \( \tau \) increases, more LoRA layers are merged, improving safety at the cost of task utility.}
  \label{fig:safemerge_threshold_analysis_gsm8k}
\end{figure}

\begin{figure*}[htbp]
  \centering

  \begin{subfigure}[t]{0.48\textwidth}
    \centering
    \includegraphics[width=\linewidth]{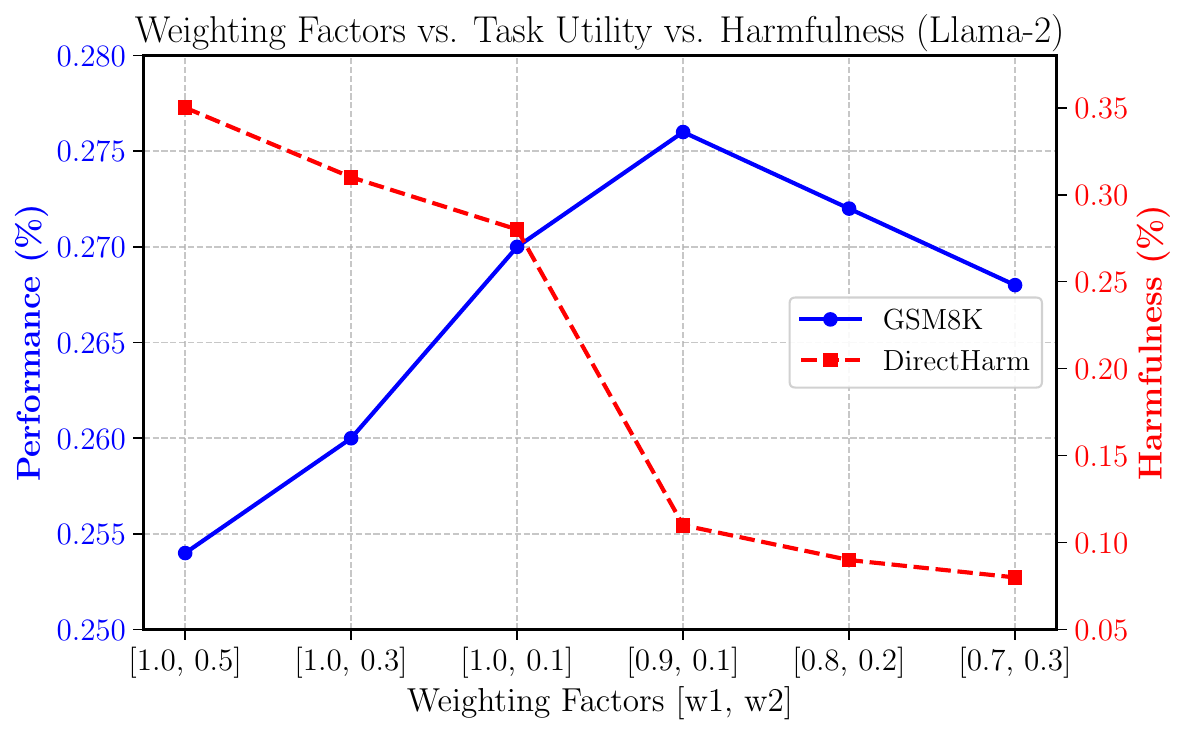}
    \caption{Llama-2-7B-Chat fine-tuned on GSM8K with \( \tau = 0.7 \).}
    \label{fig:safemerge_weighting_factor_llama_gsm8k}
  \end{subfigure}
  \hfill
  \begin{subfigure}[t]{0.48\textwidth}
    \centering
    \includegraphics[width=\linewidth]{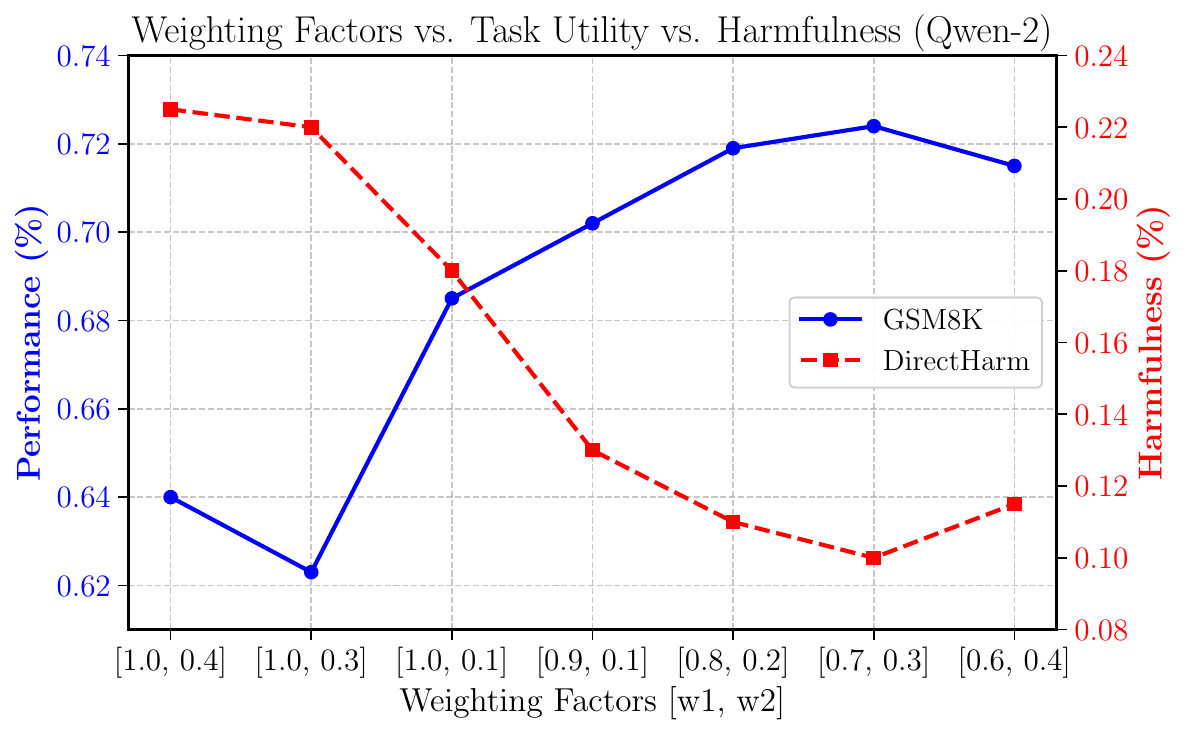}
    \caption{Qwen-2-7B-Instruct fine-tuned on GSM8K with \( \tau = 0.65 \).}
    \label{fig:safemerge_weighting_factor_qwen_gsm8k}
  \end{subfigure}

  \caption{Trade-off analysis between task utility (GSM8K) and safety (DirectHarm) for Llama-2 (left) and Qwen-2 (right), evaluated across different weighting factors \( \alpha \). Increasing the contribution of the safety-aligned model improves safety but reduces task performance. The best results are achieved when the weights sum to 1.0 during linear merging, i.e., \( \alpha \in (0, 1) \), particularly within the range of $[0.9, 0.1]$ to $[0.6, 0.4]$.} 
  \label{fig:safemerge_weighting_factors}
  
\end{figure*}

\section{SafeMERGE Results and Ablations}
\label{sec:E_results}

This section provides detailed ablation analyses along with supplemental results.

\subsection{Impact of the Tuning Threshold}
\label{sec:E_2}
Fig.~\ref{fig:safemerge_threshold_analysis_llama2_gsm8k}--\ref{fig:safemerge_threshold_analysis_qwen2_gsm8k} illustrate how the number of merged LoRA layers, task utility, and harmfulness (Direct-Harm) vary with different cosine similarity thresholds \( \tau \) for Llama-2, Llama-3.1, and Qwen-2 models fine-tuned on GSM8K. 
As \( \tau \) increases, more LoRA layers are merged, improving safety at the cost of task utility. 
In particular, merging all LoRA layers (\( \tau = 1 \)) converges to the performance of full linear merging between the fine-tuned and safety-aligned models.  
Notably, merging as few as eight layers already leads to a substantial reduction in harmfulness across all models.  
Overall, thresholds around \( \tau = 0.7 \) yield the most balanced trade-off between task utility and safety, where SafeMERGE merges 28 layers for Llama-2, 29 for Llama-3.1, and 34 for Qwen-2/2.5.  
However, selecting the optimal threshold cannot be done in isolation, as it depends on the weighting factors discussed next.

\subsection{Weighting Factors for a Given Threshold}
\label{sec:E_3}
The weighting factor \( \alpha \) determines the relative contribution of fine-tuned and safety-aligned model parameters during linear merging in SafeMERGE.  
Apart from the similarity threshold \( \tau \), tuning each model's contribution is essential for achieving an optimal trade-off between task utility and safety.  
Smaller values of \( \alpha \) indicate a greater contribution from the safety-aligned model, thereby incorporating more safe parameters during merging.  
Consequently, the similarity threshold \( \tau \) and the weighting factor \( \alpha \) should not be analyzed in isolation. 
\refig{fig:safemerge_weighting_factor_llama_gsm8k} and \refig{fig:safemerge_weighting_factor_qwen_gsm8k} show the impact of different weighting factors \( \alpha \) for a given threshold on the Llama-2 and Qwen-2 models evaluated on GSM8K.  
We observe optimal trade-offs between safety (measured on DirectHarm) and task utility for weightings that sum up to 1.0, i.e., \( \alpha \in (0, 1) \), which is in contrast to the reported optimal weightings in RESTA. 
In general, optimal ranges lie between $[0.9, 0.1]$ and $[0.6, 0.4]$, suggesting that only a small portion of the safety-aligned model is sufficient for safe merging.
Similar results can be observed for Llama-3.1 and Qwen-2.5.

\begin{table*}[htbp]
\centering
\scriptsize
\caption{SafeMERGE performance for Linear, DARE-Linear, and TIES merging on Llama models.}
\label{tab:safemerge_merging_comparison_llama}
\resizebox{\textwidth}{!}{%
    \begin{tabular}{ccccccccccccc}
    \toprule
    & \multicolumn{6}{c}{\textbf{Llama-2-7B-Chat}} 
    & \multicolumn{6}{c}{\textbf{Llama-3.1-8B-Instruct}} \\
    \cmidrule(lr){2-7}\cmidrule(lr){8-13}
    & \multicolumn{3}{c}{GSM8K} 
    & \multicolumn{3}{c}{PubMedQA} 
    & \multicolumn{3}{c}{GSM8K}
    & \multicolumn{3}{c}{PubMedQA} \\
    \cmidrule(lr){2-4}\cmidrule(lr){5-7}\cmidrule(lr){8-10}\cmidrule(lr){11-13}
    \makecell[c]{\textbf{Merging} \\ \textbf{Strategy}}
    & DirectHarm & HexPhi & Utility
    & DirectHarm & HexPhi & Utility
    & DirectHarm & HexPhi & Utility
    & DirectHarm & HexPhi & Utility \\
    \midrule
    Linear        & 7.50 & 5.70 & 26.96 & 8.10 & 4.30 & 72.20 & 8.80 & 6.30 & 78.50 & 9.10 & 6.80 & 79.00 \\
    DARE-Linear   & 8.10 & 5.70 & 26.80 & 7.90 & 4.50 & 72.40 & 9.50 & 6.70 & 78.20 & 9.60 & 7.10 & 78.70 \\
    TIES          & 5.80 & 4.60 & 26.46 & 4.30 & 3.30 & 55.20 & 12.90 & 9.70 & 74.20 & 13.80 & 11.50 & 74.60 \\
    \midrule
    Original Model & 5.00 & 2.00 & 22.67 & 5.00 & 2.00 & 55.20 & 11.30 & 7.90 & 73.80 & 11.30 & 7.90 & 74.40 \\
    \bottomrule
    \end{tabular}
}
\end{table*}

\begin{table*}[t]
\centering
\scriptsize
\caption{SafeMERGE performance for Linear, DARE-Linear, and TIES merging on Qwen models.}
\label{tab:safemerge_merging_comparison_qwen}
\resizebox{\textwidth}{!}{%
    \begin{tabular}{ccccccccccccc}
    \toprule
    & \multicolumn{6}{c}{\textbf{Qwen-2-7B-Instruct}} 
    & \multicolumn{6}{c}{\textbf{Qwen-2.5-7B-Instruct}} \\
    \cmidrule(lr){2-7}\cmidrule(lr){8-13}
    & \multicolumn{3}{c}{\textbf{GSM8K}} 
    & \multicolumn{3}{c}{\textbf{PubMedQA}} 
    & \multicolumn{3}{c}{\textbf{GSM8K}} 
    & \multicolumn{3}{c}{\textbf{PubMedQA}} \\
    \cmidrule(lr){2-4}\cmidrule(lr){5-7}\cmidrule(lr){8-10}\cmidrule(lr){11-13}
    \makecell[c]{\textbf{Merging} \\ \textbf{Strategy}}
    & DirectHarm & HexPhi & Utility
    & DirectHarm & HexPhi & Utility
    & DirectHarm & HexPhi & Utility
    & DirectHarm & HexPhi & Utility \\
    \midrule
    Linear               & 8.20  & 7.50  & 72.90 & 8.50  & 5.90  & 80.30 & 10.10 & 10.30 & 78.40 & 11.10 & 8.70 & 79.50 \\
    DARE-Linear          & 8.30  & 7.50  & 72.60 & 8.30  & 5.30  & 79.90 & 10.30 & 10.40 & 78.10 & 11.00 & 8.90 & 79.30 \\
    TIES                 & 15.80 & 12.50 & 60.73 & 18.50 & 13.80 & 75.40 & 15.50 & 13.30 & 56.70 & 14.40 & 13.50 & 75.10 \\
    \midrule
    Original Model       & 18.20 & 11.50 & 58.38 & 18.20 & 11.50 & 73.60 & 14.20 & 13.20 & 54.60 & 14.20 & 13.20 & 73.20 \\
    \bottomrule
    \end{tabular}
}
\end{table*}

\subsection{Impact of Different Merging Strategies}
\label{sec:E_4}
We report utility and safety benchmarks in Table~\ref{tab:safemerge_merging_comparison_llama} for Llama-2 and Llama-3.1 models, and in Table~\ref{tab:safemerge_merging_comparison_qwen} for Qwen-2 and Qwen-2.5, on both GSM8K and PubMedQA tasks, comparing linear, DARE-linear, and TIES merging strategies for SafeMERGE.  
Overall, we observe that linear and DARE-linear merging yield similar outcomes, with no significant deviations between them. 
In contrast, TIES merging produces inconsistent behavior. 
For Llama-2 on GSM8K, it improves safety compared to linear and DARE-linear merging while maintaining competitive utility. 
However, in all other experiments, TIES merging degrades model performance, reverting it toward baseline levels of the original (non-fine-tuned) model and, in some cases, even increasing harmfulness.  
These findings suggest that TIES merging fails to suppress harmful directions and may inadvertently reinforce them during layer-wise merging.  
A deeper analysis of this behavior is warranted and left for future work.

\subsection{Computational Analysis}
\label{sec:computational_analysis}
SafeMERGE identifies harmful LoRA layers via simple subspace operations and selectively merges them with those layers from a safe model.
Because SafeMERGE operates exclusively on model weights after fine-tuning, it requires no additional training loops, backpropagation, or gradient projections, resulting in linear computational complexity with respect to the number of LoRA parameters.

More formally, for each LoRA layer $i$, SafeMERGE computes a cosine similarity and applies linear merging when required.
As both steps are linear in the number of LoRA parameters, the total floating-point operations (FLOPs) are given as
\begin{equation}
    \text{FLOPs} = O\!\left(\sum_{i=1}^L P_i\right) = O(P_{\text{LoRA}}),
\end{equation}
where $P_i$ denotes the number of trainable parameters in layer $i$, and $P_{\text{LoRA}}$ is the total number of LoRA parameters across all $L$ layers.

In our experiments, $P_{\text{LoRA}}$ constitutes approximately 0.5--1\% of the full model’s parameters, making the computational overhead of SafeMERGE negligible compared to the cost of LoRA fine-tuning itself.
Moreover, the safe model used for merging is \emph{task-agnostic} and \emph{trained only once}, allowing it to be reused across downstream tasks without incurring additional computational cost.

\subsection{Performance against Baselines}
\label{sec:E_1}
\begin{figure*}[htbp]
  \centering
  
  \begin{subfigure}[b]{0.46\linewidth}
    \centering
    \includegraphics[width=\linewidth]{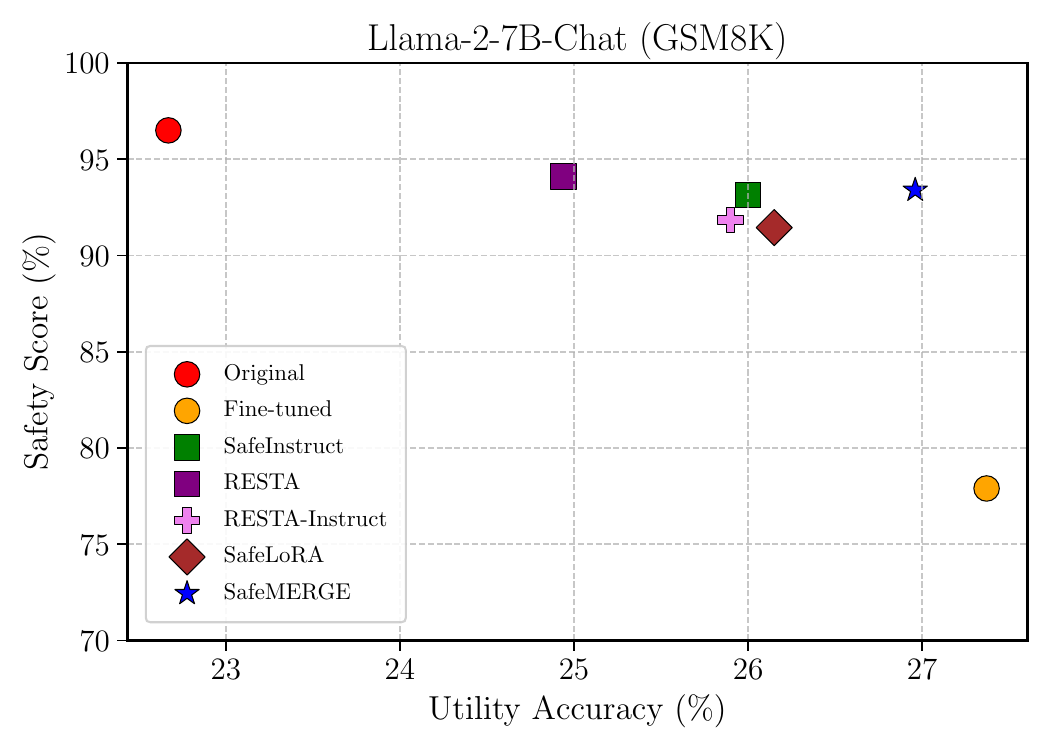}
    \caption{Llama-2-7B-Chat (GSM8K).}
    \label{fig:safemerge_comparison_llama_2_gsm8k}
  \end{subfigure}
  \hfill
  \begin{subfigure}[b]{0.46\linewidth}
    \centering
    \includegraphics[width=\linewidth]{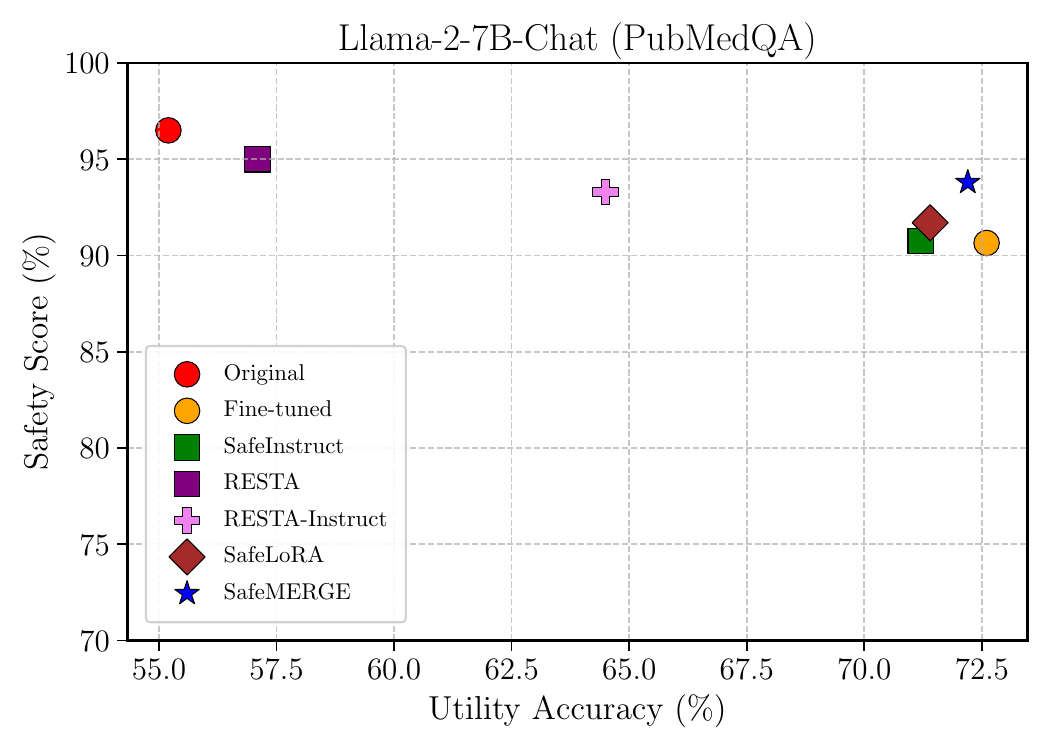}
    \caption{Llama-2-7B-Chat (PubMedQA).}
    \label{fig:safemerge_comparison_llama_2_pubmedqa}
  \end{subfigure}
  
  \vspace{1em} 

  \begin{subfigure}[b]{0.46\linewidth}
    \centering
    \includegraphics[width=\linewidth]{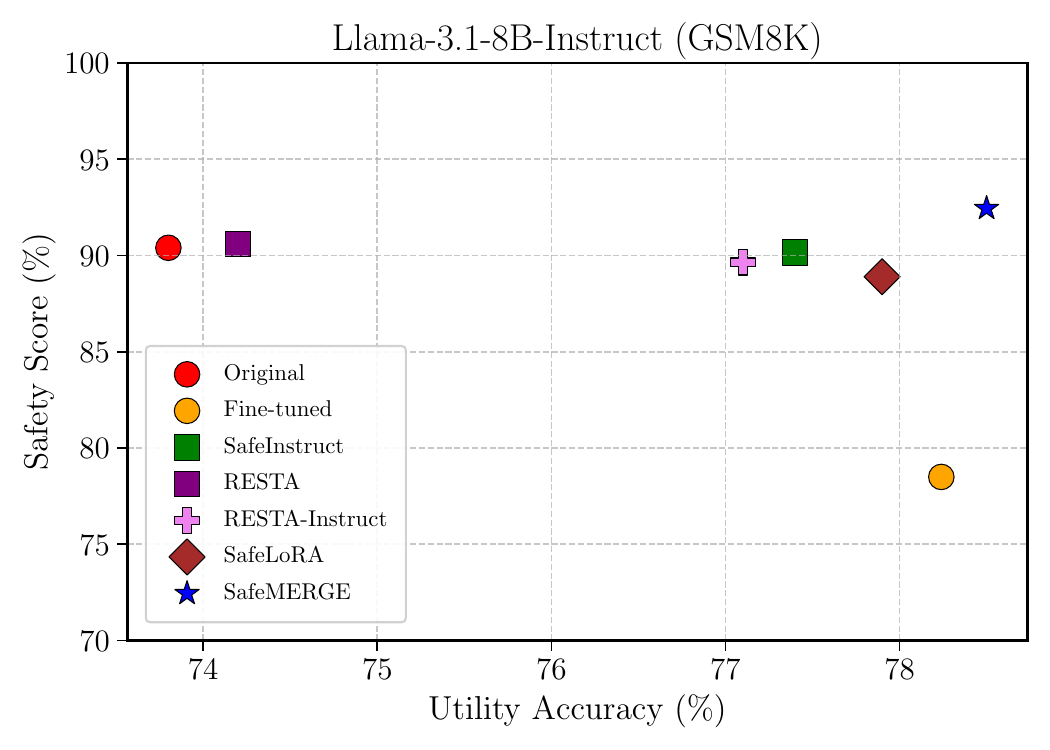}
    \caption{Llama-3.1-8B-Instruct (GSM8K).}
    \label{fig:safemerge_comparison_llama_3_gsm8k}
  \end{subfigure}
  \hfill
  \begin{subfigure}[b]{0.46\linewidth}
    \centering
    \includegraphics[width=\linewidth]{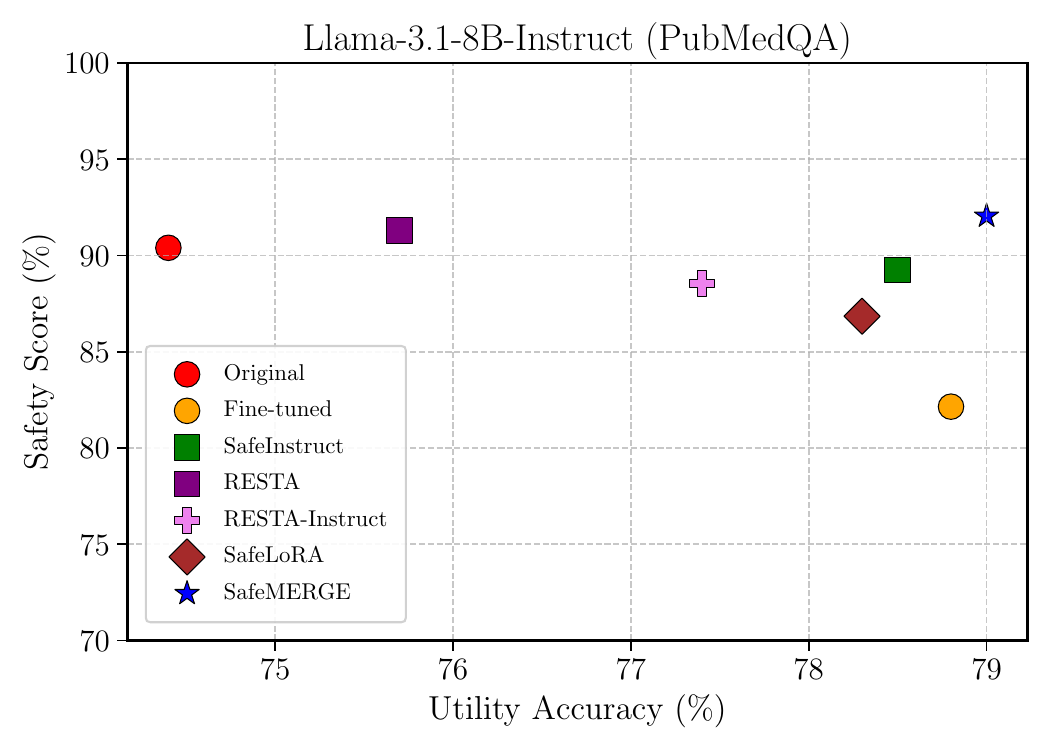}
    \caption{Llama-3.1-8B-Instruct (PubMedQA).}
    \label{fig:safemerge_comparison_llama_3_pubmedqa}
  \end{subfigure}

  \begin{subfigure}[b]{0.46\linewidth}
    \centering
    \includegraphics[width=\linewidth]{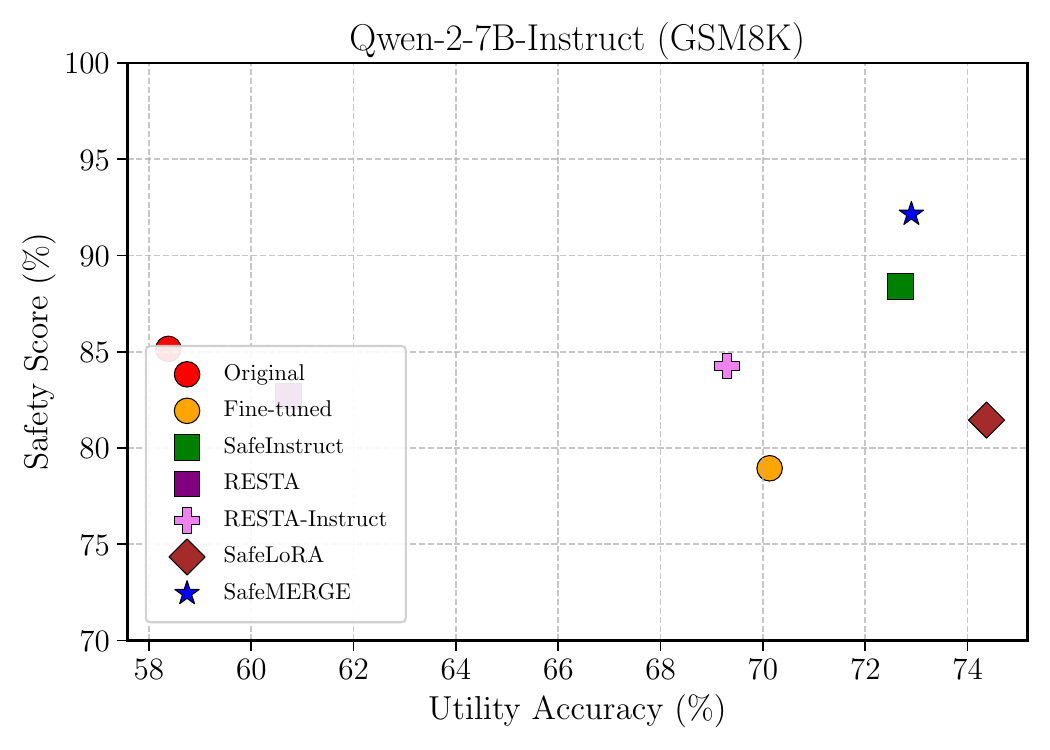}
    \caption{Qwen-2-7B-Instruct (GSM8K).}
    \label{fig:safemerge_comparison_qwen_2_gsm8k}
  \end{subfigure}
  \hfill
  \begin{subfigure}[b]{0.46\linewidth}
    \centering
    \includegraphics[width=\linewidth]{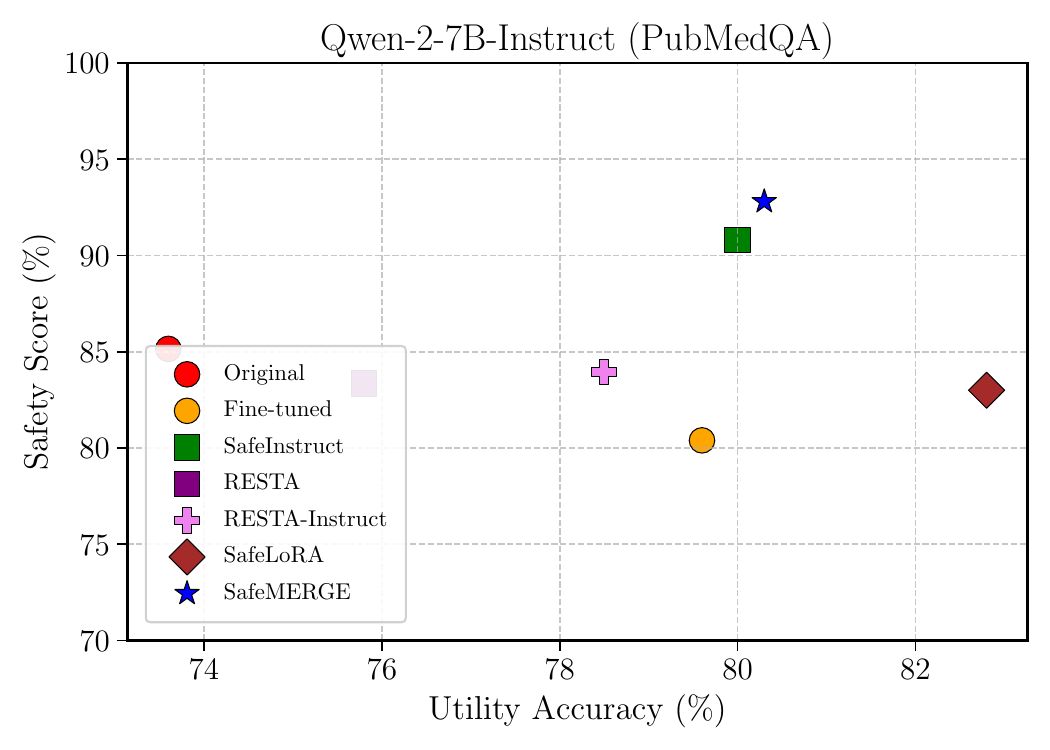}
    \caption{Qwen-2-7B-Instruct (PubMedQA).}
    \label{fig:safemerge_comparison_qwen_2_pubmedqa}
  \end{subfigure}

  \begin{subfigure}[b]{0.46\linewidth}
    \centering
    \includegraphics[width=\linewidth]{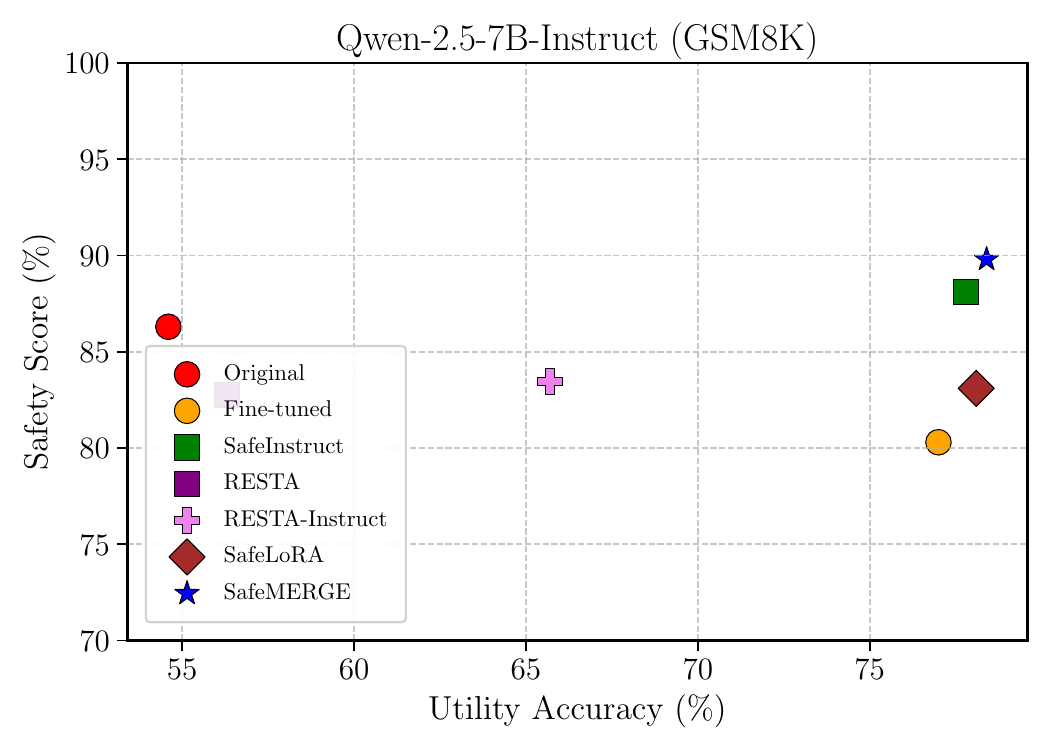}
    \caption{Qwen-2.5-7B-Instruct (GSM8K).}
    \label{fig:safemerge_comparison_qwen_2.5_gsm8k}
  \end{subfigure}
  \hfill
  \begin{subfigure}[b]{0.46\linewidth}
    \centering
    \includegraphics[width=\linewidth]{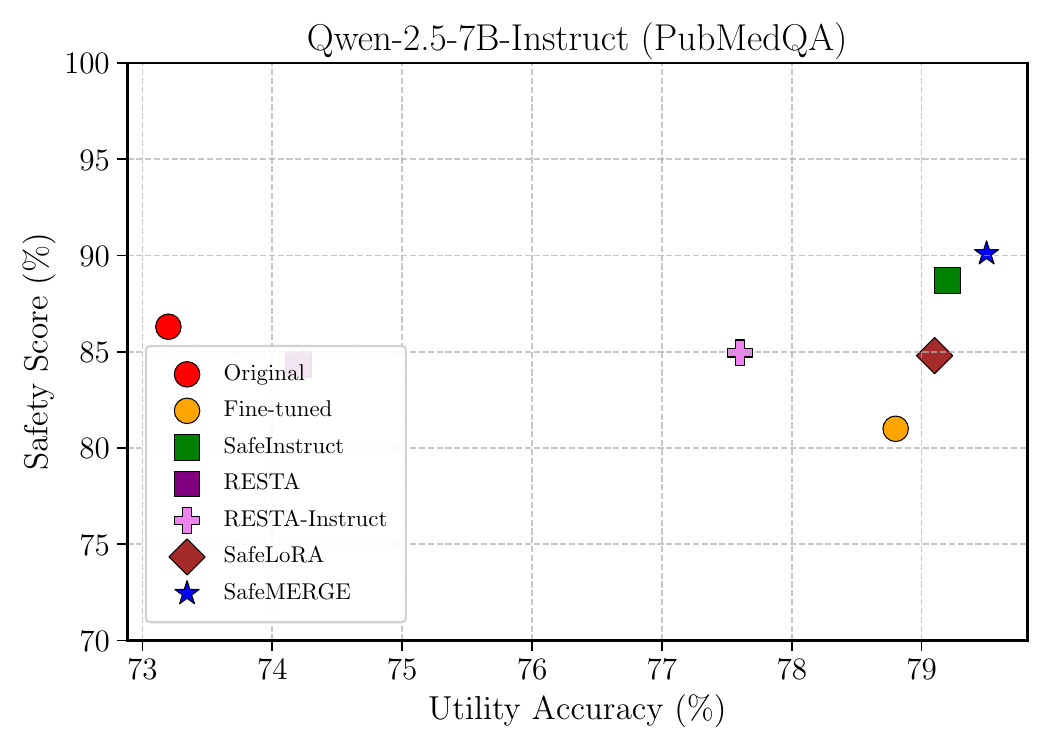}
    \caption{Qwen-2.5-7B-Instruct (PubMedQA).}
    \label{fig:safemerge_comparison_qwen_2.5_pubmedqa}
  \end{subfigure}

  \caption{SafeMERGE performance against baselines.}

  \label{fig:safemerge_scatter_plot}
\end{figure*}

\refig{fig:safemerge_scatter_plot} compares the trade-off between task utility and safety for SafeMERGE and the corresponding baselines.
To this end, we compute a compound safety score that jointly accounts for the Direct-Harm (\( d \)) and HexPhi (\( h \)) benchmarks as follows:
\begin{equation*}
    \text{Safety Score} = \frac{(100 - d) + (100 - h)}{2} .
\end{equation*}
Across all models and downstream tasks, SafeMERGE yields the best overall trade-off, achieving the highest utility with the lowest harmfulness, and thus consistently outperforming all baselines.
The results emphasize that selective, layer-wise merging is an effective strategy for maintaining safety in fine-tuned LLMs without sacrificing performance.

\section{Extended Results and Discussion}
\label{sec:extended_results}

We present additional results, ablations, and discussions that further support our main findings.

\subsection{Experiments with ShieldGemma-9B}
\label{sec:safety_guard_model}
Llama-Guard-3-8B \citep{grattafiori2024llama3} serves as the standard safety classifier in the field, following established protocols in prior work \citep{yao2024survey, qi2024safetyalignmentjusttokens, hsu2025safelorasilverlining}.
However, to address the inherent limitations of classifier-based safety assessment and mitigate potential evaluator bias, we cross-validate our results using ShieldGemma-9B \citep{shieldgemma}, a similarly sized guard model.
Table \ref{tab:safety_guard_results} presents harmfulness scores on DirectHarm and HexPhi for all defenses, as evaluated by both guard models.
We observe highly consistent trends between the two evaluators, confirming that our primary Llama-Guard-based assessment is robust and not driven by model-specific biases.

\begin{table*}[t]
    \centering
    \caption{Harmfulness scores for SafeMERGE and baselines (SafeInstruct, RESTA, RESTA-Instruct, SafeLoRA) for Llama and Qwen models, cross-validated with Llama-Guard-3B (L) and ShieldGemma-9B (G) guard models.}
    \label{tab:safety_guard_results}
    \resizebox{1\linewidth}{!}{
    \begin{tabular}{clccccccc}
        \toprule
        \textbf{Model} & \textbf{Benchmark} & \textbf{Original} & \textbf{Fine-tuned} & \textbf{SafeInstruct} & \textbf{RESTA} & \textbf{RESTA-Instruct} & \textbf{SafeLoRA} & \textbf{SafeMERGE} \\
        \midrule

        \toprule
        \multirow{2}{*}{\begin{tabular}{@{}c@{}}Llama-2-7B-Chat \\ (GSM8K)\end{tabular}}
          & DirectHarm (L/G) $(\downarrow)$ & 5.00/4.50 & 27.80/27.30 & 7.50/7.30 & 7.50/7.20 & 9.50/8.90 & 10.20/9.60 & 7.50/7.10 \\
          & HexPhi (L/G) $(\downarrow)$     & 2.00/1.50 & 16.40/16.00 & 6.20/5.90 & 4.30/4.10 & 6.80/6.30 & 6.90/6.40 & 5.70/5.20 \\
        \midrule

        \multirow{2}{*}{\begin{tabular}{@{}c@{}}Llama-2-7B-Chat \\ (PubMedQA)\end{tabular}}
          & DirectHarm (L/G) $(\downarrow)$ & 5.00/4.50 & 12.50/11.90 & 12.20/11.80 & 5.80/5.50 & 8.10/7.80 & 10.70/10.30 & 8.10/7.80 \\
          & HexPhi (L/G) $(\downarrow)$     & 2.00/1.50 & 6.20/5.90 & 6.30/5.90 & 4.20/3.80 & 5.30/5.00 & 5.90/5.60 & 4.30/4.10 \\

        \toprule
        \multirow{2}{*}{\begin{tabular}{@{}c@{}}Llama-3.1-8B-Instruct \\ (GSM8K)\end{tabular}}
          & DirectHarm (L/G) $(\downarrow)$ & 11.30/10.70 & 28.30/27.60 & 12.50/11.90 & 11.90/11.60 & 13.50/12.90 & 15.10/14.50 & 8.80/8.30 \\
          & HexPhi (L/G) $(\downarrow)$     & 7.90/7.70 & 14.70/14.30 & 7.20/6.80 & 6.90/6.70 & 7.20/6.80 & 7.10/6.80 & 6.30/6.10 \\
        \midrule

        \multirow{2}{*}{\begin{tabular}{@{}c@{}}Llama-3.1-8B-Instruct \\ (PubMedQA)\end{tabular}}
          & DirectHarm (L/G) $(\downarrow)$ & 11.30/10.70 & 23.50/23.10 & 11.80/11.30 & 10.30/9.80 & 14.20/13.80 & 16.70/16.40 & 9.10/8.80 \\
          & HexPhi (L/G) $(\downarrow)$     & 7.90/7.70 & 12.20/11.80 & 9.70/9.30 & 7.10/6.70 & 8.70/8.50 & 9.60/9.20 & 6.80/6.40 \\

        \toprule
        \multirow{2}{*}{\begin{tabular}{@{}c@{}}Qwen-2-7B-Instruct \\ (GSM8K)\end{tabular}}
          & DirectHarm (L/G) $(\downarrow)$ & 18.20/17.60 & 25.30/24.80 & 13.70/13.40 & 18.80/18.40 & 17.40/17.10 & 22.30/21.70 & 8.20/7.80 \\
          & HexPhi (L/G) $(\downarrow)$     & 11.50/11.10 & 16.80/16.10 & 9.50/9.10 & 15.80/15.30 & 14.10/13.70 & 14.80/14.20 & 7.50/6.90 \\
        \midrule

        \multirow{2}{*}{\begin{tabular}{@{}c@{}}Qwen-2-7B-Instruct \\ (PubMedQA)\end{tabular}}
          & DirectHarm (L/G) $(\downarrow)$ & 18.20/17.60 & 26.00/25.70 & 12.50/12.10 & 18.50/18.00 & 17.60/16.90 & 19.50/19.10 & 8.50/7.90 \\
          & HexPhi (L/G) $(\downarrow)$     & 11.50/11.10 & 13.20/12.80 & 5.90/5.30 & 14.80/14.40 & 14.50/13.90 & 14.50/13.90 & 5.90/5.20 \\

        \toprule
        \multirow{2}{*}{\begin{tabular}{@{}c@{}}Qwen-2.5-7B-Instruct \\ (GSM8K)\end{tabular}}
          & DirectHarm (L/G) $(\downarrow)$ & 14.20/13.70 & 22.10/21.40 & 12.50/11.80 & 18.30/17.60 & 17.70/17.10 & 19.40/18.60 & 10.10/9.60 \\
          & HexPhi (L/G) $(\downarrow)$     & 13.20/12.80 & 17.30/16.70 & 11.30/10.80 & 16.20/15.70 & 15.40/14.80 & 14.40/13.70 & 10.30/9.90 \\
        \midrule

        \multirow{2}{*}{\begin{tabular}{@{}c@{}}Qwen-2.5-7B-Instruct \\ (PubMedQA)\end{tabular}}
          & DirectHarm (L/G) $(\downarrow)$ & 14.20/13.60 & 20.20/19.40 & 13.10/12.60 & 15.70/14.40 & 14.90/14.10 & 17.10/16.40 & 11.10/10.60 \\
          & HexPhi (L/G) $(\downarrow)$     & 13.20/12.80 & 17.80/17.20 & 9.50/8.90 & 15.60/14.50 & 15.20/14.60 & 13.30/12.70 & 8.70/8.40 \\

        \bottomrule
    \end{tabular}
    }
\end{table*}

\subsection{Hyperparameter Validation on Test Sets}
\label{sec:validation_analysis}
To maintain comparability with baselines, our main experiments report results on standard test sets.
We note that widely used safety benchmarks, such as DirectHarm and HexPhi, do not provide separate training or validation splits.
Consequently, standard practice is to perform hyperparameter tuning directly on the test set.
This approach is adopted by all prior relevant work, including the baselines in our study.
Nevertheless, to address potential test set leakage, we validated SafeMERGE's hyperparameters for Llama-3.1 (GSM8K) by constructing a dedicated train-test split, using a 20\% validation set for parameter tuning and an 80\% held-out test set for final evaluation.
The search yielded an optimal configuration of \(\tau \approx 0.7\), identical to the stable default identified in our full-dataset ablations.
When evaluating the model using this validation-tuned threshold on the remaining 80\% held-out data, the safety and utility scores remained statistically consistent with our reported main results.
This confirms that SafeMERGE's hyperparameters are structural, governed by layer-wise representational drift, rather than data-specific artifacts.
Thus, our heuristics are robust to data splitting, and searching on the full set does not lead to overfitting.
We strongly advise that future red-teaming benchmarks should incorporate dedicated validation splits to facilitate distinct tuning and evaluation phases.

\subsection{Results for Telecom Domain Tasks}
\label{sec:telecom_results}

\begin{table*}[t]
\centering
\caption{SafeMERGE compared to baselines (SafeInstruct, RESTA, RESTA-Instruct, SafeLoRA) on Llama and Qwen models fine-tuned on TeleData, TeleQnA, and TSpecLLM. Utility is measured on the target task, general capabilities on IFEval/MMLU, and harmfulness via DirectHarm/HexPhi. Best results are \textbf{bolded} and second-best \underline{underlined}. MMLU and IFEval benchmarks are omitted for brevity and readibility.}
\label{tab:results_telecom}
\resizebox{1\linewidth}{!}{%
\begin{tabular}{lllccccccc}
\toprule
 & \textbf{Model} & \textbf{Benchmark} & \textbf{Original} & \textbf{Fine-tuned} & \textbf{SafeInstruct} & \textbf{RESTA} & \textbf{RESTA-Instruct} & \textbf{SafeLoRA} & \textbf{SafeMERGE} \\
\midrule

\multirow{12}{*}{\rotatebox{90}{\textbf{TeleData}}}
 & \multirow{3}{*}{\begin{tabular}{@{}c@{}}Llama-2-7B-Chat\end{tabular}}
   & TeleData $(\uparrow)$         & 29.00 & \textbf{38.70} & \textbf{38.70} & 30.10 & 32.40 & 37.30 & \underline{38.50} \\
 & & DirectHarm $(\downarrow)$     & \textbf{5.00} & 36.70 & 8.50 & 8.70 & 9.50 & 10.20 & \underline{6.90} \\
 & & HexPhi $(\downarrow)$         & \textbf{2.00} & 20.10 & 7.30 & 6.10 & 6.70 & 8.50 & \underline{5.10} \\
\cmidrule(lr){2-10}

 & \multirow{3}{*}{\begin{tabular}{@{}c@{}}Llama-3.1-8B-Instruct\end{tabular}}
   & TeleData $(\uparrow)$         & 31.70 & \textbf{47.60} & \textbf{47.60} & 34.90 & 39.50 & 46.70 & \underline{47.30} \\
 & & DirectHarm $(\downarrow)$     & 11.30 & 27.00 & \underline{10.10} & 11.70 & 13.60 & 12.70 & \textbf{8.70} \\
 & & HexPhi $(\downarrow)$         & 7.90 & 14.10 & 8.10 & \underline{6.90} & 7.40 & 8.40 & \textbf{6.10} \\
\cmidrule(lr){2-10}

 & \multirow{3}{*}{\begin{tabular}{@{}c@{}}Qwen-2-7B-Instruct\end{tabular}}
   & TeleData $(\uparrow)$         & 34.70 & \textbf{48.80} & \underline{48.70} & 36.50 & 41.90 & 46.50 & \textbf{48.80} \\
 & & DirectHarm $(\downarrow)$     & 18.20 & 34.50 & \underline{15.70} & 19.10 & 17.60 & 21.80 & \textbf{12.10} \\
 & & HexPhi $(\downarrow)$         & 11.50 & 26.30 & \underline{10.10} & 13.60 & 12.40 & 12.80 & \textbf{8.40} \\
\cmidrule(lr){2-10}

 & \multirow{3}{*}{\begin{tabular}{@{}c@{}}Qwen-2.5-7B-Instruct\end{tabular}}
   & TeleData $(\uparrow)$         & 39.80 & \underline{53.80} & 53.50 & 41.20 & 46.90 & 49.80 & \textbf{54.10} \\
 & & DirectHarm $(\downarrow)$     & \underline{14.20} & 29.70 & 14.40 & 17.80 & 17.10 & 19.70 & \textbf{11.50} \\
 & & HexPhi $(\downarrow)$         & 13.20 & 18.80 & \underline{12.90} & 16.10 & 15.40 & 16.60 & \textbf{10.40} \\

\midrule

\multirow{12}{*}{\rotatebox{90}{\textbf{TeleQnA}}}
 & \multirow{3}{*}{\begin{tabular}{@{}c@{}}Llama-2-7B-Chat\end{tabular}}
   & TeleQnA $(\uparrow)$          & 35.80 & \textbf{57.80} & 56.30 & 41.40 & 49.60 & 57.00 & \underline{57.20} \\
 & & DirectHarm $(\downarrow)$     & \textbf{5.00} & 12.30 & 6.80 & 7.70 & 6.40 & 7.50 & \underline{5.90} \\
 & & HexPhi $(\downarrow)$         & \textbf{2.00} & 7.50 & 4.20 & 5.60 & 4.50 & 5.00 & \underline{3.80} \\
\cmidrule(lr){2-10}

 & \multirow{3}{*}{\begin{tabular}{@{}c@{}}Llama-3.1-8B-Instruct\end{tabular}}
   & TeleQnA $(\uparrow)$          & 42.30 & \textbf{67.80} & 66.80 & 52.20 & 60.90 & 65.30 & \underline{67.10} \\
 & & DirectHarm $(\downarrow)$     & 11.30 & 18.20 & 9.50 & \underline{8.80} & 10.80 & 11.00 & \textbf{8.20} \\
 & & HexPhi $(\downarrow)$         & 7.90 & 11.80 & 6.20 & \underline{6.00} & 6.80 & 7.10 & \textbf{5.80} \\
\cmidrule(lr){2-10}

 & \multirow{3}{*}{\begin{tabular}{@{}c@{}}Qwen-2-7B-Instruct\end{tabular}}
   & TeleQnA $(\uparrow)$          & 45.80 & \textbf{65.60} & 64.80 & 49.80 & 60.40 & 64.10 & \underline{65.20} \\
 & & DirectHarm $(\downarrow)$     & 18.20 & 26.30 & \underline{13.70} & 19.10 & 16.80 & 19.20 & \textbf{11.80} \\
 & & HexPhi $(\downarrow)$         & 11.50 & 15.80 & \underline{8.50} & 13.10 & 11.90 & 11.30 & \textbf{7.50} \\
\cmidrule(lr){2-10}

 & \multirow{3}{*}{\begin{tabular}{@{}c@{}}Qwen-2.5-7B-Instruct\end{tabular}}
   & TeleQnA $(\uparrow)$          & 48.70 & 69.90 & \textbf{70.10} & 50.60 & 60.20 & 68.40 & \underline{70.00} \\
 & & DirectHarm $(\downarrow)$     & \underline{14.20} & 21.20 & 15.10 & 17.00 & 16.30 & 17.40 & \textbf{10.90} \\
 & & HexPhi $(\downarrow)$         & 13.20 & 16.10 & \underline{12.80} & 15.80 & 15.20 & 14.90 & \textbf{9.60} \\

\midrule

\multirow{12}{*}{\rotatebox{90}{\textbf{TSpecLLM}}}
 & \multirow{3}{*}{\begin{tabular}{@{}c@{}}Llama-2-7B-Chat\end{tabular}}
   & TSpecLLM $(\uparrow)$         & 33.30 & \textbf{44.20} & \underline{43.90} & 35.80 & 39.20 & 42.90 & 43.80 \\
 & & DirectHarm $(\downarrow)$     & \textbf{5.00} & 12.90 & 7.50 & 6.80 & 7.90 & 8.20 & \underline{6.30} \\
 & & HexPhi $(\downarrow)$         & \textbf{2.00} & 7.30 & 4.90 & 4.70 & 5.40 & 6.40 & \underline{4.50} \\
\cmidrule(lr){2-10}

 & \multirow{3}{*}{\begin{tabular}{@{}c@{}}Llama-3.1-8B-Instruct\end{tabular}}
   & TSpecLLM $(\uparrow)$         & 48.50 & \textbf{62.10} & 61.50 & 51.70 & 58.90 & 60.80 & \underline{61.90} \\
 & & DirectHarm $(\downarrow)$     & 11.30 & 17.50 & 9.80 & \underline{9.40} & 11.20 & 11.40 & \textbf{8.50} \\
 & & HexPhi $(\downarrow)$         & 7.90 & 10.70 & 5.90 & \underline{5.70} & 6.60 & 7.30 & \textbf{5.10} \\
\cmidrule(lr){2-10}

 & \multirow{3}{*}{\begin{tabular}{@{}c@{}}Qwen-2-7B-Instruct\end{tabular}}
   & TSpecLLM $(\uparrow)$         & 12.50 & \textbf{28.30} & 28.00 & 13.80 & 24.50 & 27.70 & \underline{28.10} \\
 & & DirectHarm $(\downarrow)$     & 18.20 & 26.60 & \underline{14.80} & 19.00 & 18.40 & 18.30 & \textbf{12.60} \\
 & & HexPhi $(\downarrow)$         & 11.50 & 16.10 & \underline{9.70} & 13.40 & 12.10 & 12.30 & \textbf{8.60} \\
\cmidrule(lr){2-10}

 & \multirow{3}{*}{\begin{tabular}{@{}c@{}}Qwen-2.5-7B-Instruct\end{tabular}}
   & TSpecLLM $(\uparrow)$         & 32.50 & \underline{52.80} & 51.50 & 34.00 & 45.80 & 49.70 & \textbf{53.10} \\
 & & DirectHarm $(\downarrow)$     & \underline{14.20} & 22.50 & \textbf{14.10} & 18.00 & 17.60 & 17.20 & \textbf{10.10} \\
 & & HexPhi $(\downarrow)$         & \underline{13.20} & 18.90 & 13.50 & 15.70 & 16.10 & 16.10 & \textbf{9.90} \\

\bottomrule
\end{tabular}
}
\end{table*}

We provide extended results for three additional utility benchmarks from the telecom domain: TeleData \citep{TeleLLMs}, TeleQnA \citep{TeleQnA}, and TSpecLLM \citep{TSpecLLM}.
These datasets contain various telecom-specific questions drawn from standards, implementations, and engineering practice, often formatted as lists, tables, and complex mathematical formulas, which are shown to be harmful during training \citep{he2024safedataidentifyingbenign, djuhera2026safecommstudysafetydegradation}.
For all datasets, we create 80/20 train-test splits and fine-tune each model using the same LoRA settings from Table~\ref{tab:lorasettings} with an effective batch size of 32 and a learning rate of 1e-4 with linear scheduling. 
We train for 2 epochs on TeleData and for 5 epochs on TeleQnA and TSpecLLM.
We then evaluate the model performance on the test split by following the approach in \citet{TeleLLMs} where we use Mixtral-8x7B-Instruct~\citep{jiang2024mixtralexperts} as a judge to compare answers with ground truth responses.
We compute the final accuracy as the ratio of correctly answered questions and assess safety as in \refapp{sec:C_2}.

In Table~\ref{tab:results_telecom}, we compare SafeMERGE against SafeInstruct and SafeLoRA. 
For SafeInstruct, we interleave a subset of harmful QA pairs (with safe refusals) from \citet{bianchi2024safetytunedllamaslessonsimproving} into the fine-tuning sets. Specifically, we inject 2500, 1000, and 10 safety samples into TeleData, TeleQnA, and TSpecLLM datasets, respectively.
For SafeLoRA, we define the safety-aligned subspace using the respective instruct/chat and base models. We choose the same optimal cosine similarity thresholds of 0.7, 0.75, 0.65, and 0.7 for Llama-2, Llama-3.1, Qwen-2, and Qwen-2.5 models, respectively.
For SafeMERGE, we follow the same procedure and additionally apply linear merging with \( \alpha = 0.7 \) across models. The safe reference model used for merging is obtained by fine-tuning each LLM on 1000 samples from \citet{bianchi2024safetytunedllamaslessonsimproving}.

In general, the results confirm previous trends.
SafeInstruct, SafeLoRA, and SafeMERGE can successfully restore safety while preserving utility.
Overall, SafeMERGE provides the best trade-off between utility and safety, followed by SafeInstruct.
For Llama-3.1, Qwen-2, and Qwen-2.5, harmfulness can, in most cases, be reduced even below that of the original instruct models.
These results confirm the effectiveness of SafeMERGE in a highly specialized domain, confirming its broad generalizability.

\end{document}